\documentclass{article} 
\usepackage{collas2026_conference,times}
\usepackage{easyReview}

\usepackage{algorithm}
\usepackage{algpseudocode}
\usepackage{setspace}
\usepackage{enumitem}
\usepackage{xcolor}

\usepackage{graphicx}
\usepackage{booktabs}
\usepackage{caption}
\usepackage{subcaption}

\usepackage{booktabs}    
\usepackage{multirow}    
\usepackage{graphicx}    

\newtheorem{lemma}{Lemma}

\floatname{algorithm}{Algorithm}

\newcommand{\vecw}{\mathbf{w}}

\newcommand{\blfootnote}[1]{%
  \begingroup
  \renewcommand{\thefootnote}{\fnsymbol{footnote}}%
  \footnotetext[0]{#1}%
  \endgroup
}

\usepackage{amsmath,amsfonts,bm}

\def\eqref#1{equation~\ref{#1}}

\def\1{\bm{1}}

\DeclareMathAlphabet{\mathsfit}{\encodingdefault}{\sfdefault}{m}{sl}
\SetMathAlphabet{\mathsfit}{bold}{\encodingdefault}{\sfdefault}{bx}{n}

\usepackage{hyperref}
\hypersetup{
    colorlinks=true,
    linkcolor=red,
    filecolor=magenta,
    urlcolor=blue,
    citecolor=purple,
    pdftitle={Overleaf Example},
    pdfpagemode=FullScreen,
    }

\title{Federated Continual Learning as a Distributed Drift-Plus-Penalty Control Problem}

\author{Nazreen Shah \\
IIIT Delhi\\
India \\
\texttt{nazreens@iiitd.ac.in} \\
\And 
Naveen Kumar Reddy Somireddy  \\
IIIT Delhi \\
India \\
\texttt{naveens@iiitd.ac.in}
\And 
Zubair Shaban \\
IIIT Delhi \\
India \\
\texttt{zubairs@iiitd.ac.in}
\AND
\quad \quad \quad \quad \quad  \quad \quad Ranjitha Prasad \\
\quad \quad \quad \quad \quad \quad \quad IIIT Delhi \\
\quad \quad \quad \quad \quad \quad \quad India \\
\quad \quad \quad \quad \quad \quad \quad \texttt{ranjitha@iiitd.ac.in}
\And
B. N. Bharath  \\
IIT Dharwad \\
India \\
\texttt{bharathbn@iitdh.ac.in}
}

\collasfinalcopy 

\begin{document}

\maketitle

\begin{abstract}
Federated Continual Learning (FCL) is fundamental to real-world distributed learning systems, requiring models to adapt to sequential, non-IID data across clients while mitigating catastrophic forgetting and client drift. Existing approaches formulate continual learning (CL) as a sequence of per-task optimization problems, applied locally at each client and coupled through aggregation, using heuristic mechanisms such as replay, regularization, or projection-based constraints. However, forgetting in FCL is inherently a long-term, distributed phenomenon, arising from the interaction of temporal task evolution and cross-client heterogeneity, which is not explicitly regulated. In this work, we cast FCL as a stochastic control problem and propose \emph{Federated Queue-regulated Continual Learning (\textsc{FedQCL})}, a framework based on Lyapunov drift-plus-penalty (DPP) optimization. \textsc{FedQCL} introduces virtual queues to track the accumulation of forgetting across tasks and clients, enabling explicit control of the stability-plasticity trade-off. By optimizing a DPP objective, the method jointly improves current-task performance while the queue-based formulation provides an interpretable and tunable mechanism to balance adaptation and retention through a single parameter, without requiring gradient projection or additional communication overhead.  Empirical evaluations on standard benchmarks, including Split-CIFAR-10, Split-CIFAR-100, and Split-TinyImageNet, demonstrate that \textsc{FedQCL} outperforms state-of-the-art baselines with respect to accuracy while significantly reducing forgetting under heterogeneous data distributions.
\end{abstract}
\section{Introduction}
\label{sec:intro}

In many real-world scenarios, such as mobile and wearable systems, data arrives sequentially, and it is essential to learn from streaming, non-IID (non-independent and identically distributed) data generated across decentralized and privacy-constrained devices. Federated Learning (FL)~\cite{FedAvg} has emerged as a privacy-preserving paradigm for training machine learning (ML) models across distributed clients without sharing raw data. However, most existing FL approaches assume static local datasets and a single task per client. In real-world applications, tasks evolve over time, requiring models to be continually updated. This necessitates a Continual Learning (CL) framework, where models must adapt to new streaming tasks while retaining knowledge of previously learned ones. In distributed settings, this gives rise to Federated Continual Learning (FCL), which combines the privacy-preserving properties of FL with the sequential adaptation requirements of CL. A central challenge in this setting is catastrophic forgetting, defined as the phenomenon where a model rapidly loses previously acquired knowledge when trained on new tasks~\cite{mccloskey1989catastrophic}. This issue is further exacerbated in federated environments due to statistical heterogeneity across clients where non-IID data distributions induce divergent local updates, leading to instability in the global learning dynamics and amplifying the forgetting of past tasks \cite{dupuy2023quantifying}.

 Regularization-based methods in FCL mitigate catastrophic forgetting by constraining parameter updates, without requiring raw data storage,  thus adhering to federated privacy constraints~\cite{Fedweit,li2024personalized,lee2024fedsol}. However, this reliance on parameter-level constraints introduces inherent limitations as constraints compound over tasks and progressively restrict the feasible update space, limiting plasticity in dynamic environments. Generation-based methods provide a data-free way to approximate past distributions, i.e., instead of storing or sharing real data, each client learns a generative model that can recreate past samples ~\cite{qibetter2023,zhang2023target,liang2024diffusion}. However, these methods incur significant computational and communication overheads. Gradient projection–based FCL methods~\cite{saha2021gradient}, such as FOT~\cite{fot}, attempt to alleviate these limitations by projecting current-task gradients onto subspaces that preserve previously learned knowledge, often facilitated through the exchange of high-dimensional embeddings across clients; however, this incurs significant communication overhead and introduces additional privacy concerns. Replay-based methods in FCL mitigate forgetting by augmenting current-task updates with samples drawn from a replay buffer that stores past data. In centralized settings, this buffer is shared across all data and is highly effective at stabilizing training and preserving prior knowledge~\cite{ER}. However, such centralized replay violates federated privacy constraints, and therefore FCL methods rely on local replay, where each client maintains its own private replay buffer constructed from its historical data~\cite{dong2022federated,liu2023fedet,li2024sr}.

In FCL, ensuring the stability of the global learning dynamics becomes as critical as achieving strong instantaneous performance in each round. One of the key limitations of all existing formulations is that while FCL is inherently a dynamic, distributed process with temporally and cross-client coupled objectives, most approaches optimize it through myopic, round-wise updates that do not explicitly account for the accumulation of forgetting over time. This motivates a control-theoretic formulation of FCL where instead of a purely optimization-based view, we adopt tools from stochastic control, in particular Lyapunov analysis and the Drift-Plus-Penalty (DPP) framework \cite{neely2006energy}, to model and regulate the dynamics of forgetting under communication and resource constraints. The DPP framework enables a principled treatment of FCL by jointly accounting for instantaneous learning performance (plasticity) and the accumulation of forgetting (stability) across clients and rounds. By explicitly incorporating both short-term objectives and long-term stability considerations, it provides a systematic mechanism to balance plasticity and stability in distributed, non-IID environments.

\textbf{Contributions:}~We cast FCL as a long-term constrained stochastic control problem, where catastrophic forgetting manifests as the accumulation of constraint violations over time and across clients. We propose \emph{Federated Queue-regulated Continual Learning (\textsc{FedQCL})}, a principled framework based on Lyapunov optimization. The novel approach in \textsc{FedQCL} is the introduction of a virtual queue that tracks the evolution of constraint violations, enabling explicit monitoring of forgetting across communication rounds and clients. At each round, clients compute local updates guided by both the immediate learning objective and the state of this queue. These updates are aggregated by the server, leading to the balancing of plasticity with global stability. The key features of our framework is as follows:
\begin{itemize}
\item \textbf{Single-parameter control of stability–plasticity trade-off:} \textsc{FedQCL} achieves a principled trade-off between adaptation to new tasks and retention of past knowledge through a single hyperparameter, which balances the current task learning objective against accumulated forgetting.

\item \textbf{Fine-grained and communication-efficient regulation:} We show that tracking and updating virtual queues at the level of communication rounds instead of task boundaries is critical, as forgetting evolves continuously due to iterative updates and partial participation. Importantly, these queues are maintained locally at each client, preserving privacy and no additional communication overhead.

\item \textbf{Empirical evaluation:} We demonstrate that \textsc{FedQCL} outperforms state-of-the-art baselines with respect to average accuracy and forgetting under heterogeneous and non-stationary data distributions.
\end{itemize}

The remainder of the paper is organized as follows: we first review related work, then present the problem formulation, followed by the proposed method, and finally provide experimental results and discussion. \blfootnote{Code available at: \url{https://github.com/Naveensomireddy4/FedQCL_Task}}

\section{Related Works}
\label{sec:rel_works}
\textbf{Classical FL and Replay-based CL methods}: FL aims to train a global model across distributed clients without sharing raw data. The {FedAvg} algorithm \cite{FedAvg} performs iterative rounds of local stochastic gradient descent (SGD) at clients followed by weighted averaging at the server. While simple and communication-efficient, {FedAvg} suffers under statistical heterogeneity, leading to client drift and degraded convergence. To address this, {FedProx} \cite{li2020federated} introduces a proximal term to limit deviation from the global model, while {SCAFFOLD} \cite{karimireddy2020scaffold} uses control variates to correct client drift, both improving stability under heterogeneity. Recent works continue to build upon these classical foundations by addressing scalability, heterogeneity, and robustness challenges~\cite{kairouz2021advances,li2023federated,wang2024federated}. Despite these advances, classical FL methods are primarily designed for static data distributions and single-task settings, and they are not suitable when data and tasks evolve over time.
Rehearsal-based CL approaches mitigate catastrophic forgetting by maintaining a limited memory buffer of past samples and augmenting current updates with replayed data. Experience Replay (ER) \cite{ER} forms the canonical baseline, with extensions such as {DER}/{DER++} \cite{der} (output distillation), {MER} \cite{mer} (meta-learning), and {ER-ACE} (asymmetric losses) improving stability. Other methods include {GDumb} \cite{gdumb}, which relies on buffer-based retraining, {CBA} \cite{cba}, which mitigates recency bias, and {REFRESH} \cite{refresh}, which adopts an unlearn–relearn strategy. Recent works refine buffer construction and allocation strategies \cite{zhang2024core}. Despite their effectiveness, these methods primarily regularize updates implicitly through replay and do not explicitly model the temporal accumulation of forgetting.

\textbf{Federated Continual Learning Methods:}
FCL extends federated learning to non-stationary environments where client data evolves over time, requiring models to retain prior knowledge while adapting to new tasks. Existing approaches combine continual learning strategies with federated optimization. Replay-based methods leverage local buffers to revisit past data and mitigate forgetting~\cite{dong2022federated,liu2023fedet,li2024sr}, while regularization and distillation-based techniques constrain updates to preserve previously acquired knowledge~\cite{lee2024fedsol}. More recent methods explicitly account for both data heterogeneity and sequential task learning. For instance, \textbf{BI}~\cite{Biserra2025federated} employs uncertainty-aware memory management to prioritize informative samples for replay, whereas \textbf{FOT}~\cite{fot} mitigates global catastrophic forgetting by projecting local updates onto the orthogonal complement of subspaces spanned by past-task gradients. In the federated setting, \cite{pmlr-v258-keshri25a} proposes C-FLAG, a replay-based CFL strategy with convergence guarantees. Despite these advances, most existing FCL methods optimize per-round objectives and do not explicitly model the accumulation of forgetting over time. {Other recent works address specific systems-level challenges in FCL:~\cite{shenaj2023asynchronous} tackle asynchronous client participation,~\cite{wang2024traceable} focus on traceability of task contributions,~\cite{wuerkaixi2024accurate} address accurate forgetting under heterogeneous distributions, and~\cite{li2026resource} consider resource-constrained settings.}

\textbf{Drift-Plus-Penalty in Machine Learning:}~ The DPP technique, first introduced in the context of stochastic network optimization ~\cite{neely2010stochastic}, is a powerful framework to design stochastic control algorithms. DPP is based on Lyapunov optimization that balances two competing objectives: minimizing a time-averaged penalty and ensuring constraint satisfaction through the control of virtual queues that are based on time-averaged constraint violations. The DPP framework has found broad applications across domains such as wireless resource allocation~\cite{neely2010efficient}, energy-efficient communications~\cite{samarakoon2015energy}, and dynamic scheduling, where it ensures online decision-making with provable convergence bounds. Recently, DPP has been used for constrained reinforcement learning, where the algorithm balances reward maximization with long-term constraint satisfaction \cite{huang2021multi,xu2025lyapunov}. {The authors in~\cite{wang2023federatedinfocom} apply the drift-plus-penalty framework to stationary federated learning settings for communication scheduling, where a single queue tracks gradient divergence}. In online learning settings, it enables dynamic regularization and adaptive constraint enforcement, leading to the best known static regret bounds \cite{sinha2024optimal, sarkar2025projection, vaze2025sqrt}. Recently, \cite{shah2026theoretical} applies DPP to replay-based CL via virtual queues, but in a centralized single-learner setting that ignores client drift and partial participation.

\textbf{Novelty:}~Overall, prior work addresses either forgetting (CL) or heterogeneity (FL), but optimizes per-round objectives and does not explicitly model the accumulation of forgetting over time. This is a gap that \textsc{FedQCL} aims to fill. As compared to many existing algorithms, \textsc{FedQCL} avoids gradient projection or dual-update mechanisms. Instead, constraints are enforced implicitly through virtual queue dynamics at the client level, enabling projection-free local updates, preserving plasticity and retaining privacy. Further, the proposed approach works under a fixed learning rate across clients and rounds, simplifying implementation and distinguishing \textsc{FedQCL} from approaches that rely on adaptive or client-specific step sizes. To the best of the authors' knowledge, \textsc{FedQCL} is the first to cast FCL as a queue-stabilization problem and explicitly model forgetting as a controllable, accumulative process across time and clients, leading to explicit control on the stability–plasticity trade-off in distributed non-IID settings.

\section{Problem Formulation}
\label{sec:prbFrm}

We consider the problem of continually fine-tuning a global model on distributed streaming data, where the model exhibits task-specific adaptations over time. Towards this, we assume a federated learning system which consists of $N$ clients. In a FCL setting, each client $i \in [N]$ observes a sequence of tasks $\{\mathcal{D}^{(t)}_i\}_{t = 1}^{T}$  and at each task $t$, the goal is to learn a global model $\vecw^{(t)}$ by aggregating contributions from all clients. If one solves each task in isolation, the corresponding standard per-task federated objective for the $t$-th task is given as:
\begin{align}
    \min_{\vecw^{(t)} \in \mathbb{R}^d} \; \Phi(\vecw^{(t)}) 
    := \sum_{i=1}^{N} p^{(t)}_i \, \Phi_i\big(\vecw^{(t)}; \mathcal{D}_i^{(t)}\big),
\end{align}
where $\Phi_i\big(\vecw^{(t)}; \mathcal{D}_i^{(t)}\big)$ denotes the local loss at client $i$ for task $t$, and $p^{(t)}_i$ are client-specific per-task weights. While the per-task formulation captures the objective at a fixed time step, it treats each task independently and does not account for the temporal evolution of data across tasks. In realistic federated settings, however, data distributions are non-stationary, and models must continually adapt to new tasks while retaining knowledge from previous ones. Solving each task in isolation can therefore lead to suboptimal performance due to issues such as catastrophic forgetting and lack of knowledge transfer. To address these challenges, we move beyond independent per-task optimization and consider a FCL formulation where the clients collaboratively solve the sequential distributed optimization problem given as:
\begin{align}
    \min_{\vecw^{(1)},\hdots,\vecw^{(t)}} {\textstyle\sum}_{t = 1}^T{\textstyle\sum}_{i = 1}^N p^{(t)}_i \Phi_i(\vecw^{(t)};\mathcal{D}^{(t)}_i).
\end{align}
An important limitation in CL when learning a new task is that data from earlier tasks is inaccessible. As a consequence, although the objective is written over all tasks, in practice the optimization is performed sequentially, with $\vecw^{(t)}$ updated using the current task data $\mathcal{D}^{(t)}_i$ and relies on surrogate mechanisms to preserve previously acquired knowledge.  We assume synchronous task boundaries across clients, i.e., all clients transition between tasks simultaneously, while  their data distributions  differs. 

The non-IID nature of client data induces statistical heterogeneity, leading to conflicting updates and exacerbating the stability–plasticity trade-off during FCL. The goal is to continually adapt the global model to new tasks while retaining past knowledge, and to effectively aggregate non-IID client updates to learn a stable and consistent global direction. To mitigate catastrophic forgetting in FCL, replay buffers are commonly employed to retain a subset of data from previously observed tasks. To maintain the privacy constraints in federated settings, each client maintains a \emph{local} buffer that stores representative samples from past tasks, which are used to approximate the loss on past data. These buffers enable the evaluation of past-task performance during current updates, thereby providing a mechanism to regulate forgetting. In particular, in our framework $\hat{\Phi}^{(k)}_{i}(\cdot)$ denotes the replay loss evaluated at client $i$ on task $k < t$. Here, the subscript $i$ indexes the client, while $k$ indexes the past task. Thus, $\hat{\Phi}^{(k)}_{i}$ captures the performance of a model on the data corresponding to the $k$-th task as approximated using the local replay buffer at client $i$. With the replay-mechanism in place, we consider the following problem where, at task $t$, the goal is to minimize the average loss across tasks while constraining the average forgetting across clients:
\begin{align} \label{eq:cfl_dpp}
    &\min_{\vecw^{(1)},\ldots,\vecw^{(t)}} 
    && \frac{1}{t}\sum_{\tau=1}^t \sum_{i=1}^N p^{(\tau)}_i \, \Phi_i(\vecw^{(\tau)};\mathcal{D}_i^{(\tau)}) \nonumber \\
    &\text{subject to} 
    && \frac{1}{t-1}\sum_{k=1}^{t-1} \sum_{i=1}^N p^{(k)}_i 
    \left(\hat{\Phi}^{(k)}_{i}(\vecw^{(t)}) - \hat{\Phi}^{(k)}_{i}(\vecw^{(t-1)})\right) \leq \delta,
\end{align}
where $\vecw^{(t-1)}$ denotes the global model obtained at the previous task and serves as a reference for measuring forgetting. The constraint enforces that the average increase in replay loss across past tasks and clients remains bounded by a tolerance $\delta$, ensuring controlled forgetting under non-IID data distributions.

\section{Proposed Method: Drift-Plus-Penalty Formulation}

To handle the constraints in the federated optimization problem formulated in \eqref{eq:cfl_dpp}, we adopt a Lyapunov optimization framework, a technique widely used for controlling constraint violations in stochastic and distributed systems~\cite{neely2013dynamic}. In this setting, each client locally minimizes a drift-plus-penalty (DPP) objective that combines the loss on the current task with a drift term capturing violations of forgetting constraints across past tasks. While the constraint in \eqref{eq:cfl_dpp} is defined as a global average over clients and tasks, directly enforcing such a coupled constraint is impractical in federated settings due to partial participation and decentralized data access. To obtain a tractable and distributed implementation, we decompose the global constraint into per-client, per-task components and track their evolution locally. Specifically, inspired by Lyapunov drift-plus-penalty method, we introduce a virtual queue $Q^{(k)}_{i}[t]$ at each client $i$ for every past task $k < t$, which accumulates the violation of the corresponding forgetting constraint over time. This yields the following update:
\begin{equation}
    Q^{(k)}_{i}[t] = \max \left\{ Q^{(k)}_{i}[t-1] 
    + \left[\hat{\Phi}^{(k)}_{i}(\vecw_i^{(t)}) - \hat{\Phi}^{(k)}_{i}(\vecw^{(t-1)}) - \delta\right], \, 0 \right\},
    \label{eq:queue_update}
\end{equation}
where $\hat{\Phi}^{(k)}_{i}(\cdot)$ denotes the local replay loss at client $i$ on task $k$, and $\delta$ is the allowable forgetting threshold. In the above expression, the queue evolves based on the instantaneous increase in replay loss at client $i$, which is a localized measure of forgetting. This decomposition enables distributed tracking of constraint violations under partial participation, while remaining consistent with the global objective in expectation. 

For each client $i$, the local Lyapunov function is defined as:
\begin{equation}
    L_i[t] = \frac{1}{2} \sum_{k=1}^{t-1} (Q^{(k)}_{i}[t])^2,
\end{equation}
with drift given by $\Delta L_i[t] = L_i[t] - L_i[t-1]$. At task $t$, each client minimizes a DPP objective, {where $V > 0 $ is a scaling parameter that governs the stability-plasticity trade-off by controlling the relative weight of the current-task loss relative to the accumulated forgetting penalty.}:

\begin{equation}
    \texttt{DPP}^{(t)}_{V,i} := \min_{\mathbf{w}^{(t)}} 
    \Big[
        V\, \Phi_i\big(\mathbf{w}^{(t)}; \mathcal{D}_i^{(t)}\big) 
        + \Delta L_i[t]
    \Big].
    \label{eq:dpp_1}
\end{equation}

The following lemma provides an upper bound on the per-client Lyapunov drift in terms of the deviation in replay losses across past tasks. 

\begin{lemma}
\label{lem:client_drift_bound_unnormalized}
For each client $i \in [N]$, the Lyapunov drift at task $t$ satisfies the following upper bound for any model $\mathbf{w} \in \mathbb{R}^d$:
\begin{align}
    \Delta L_i[t] 
    \leq \frac{1}{2} \sum_{k=1}^{t-1} 
    \left( \Delta \hat{\Phi}^{(k)}_{i}(\mathbf{w}, \mathbf{w}^{(t-1)}) \right)^2
    + \sum_{k=1}^{t-1} Q^{(k)}_{i}[t-1] \, \Delta \hat{\Phi}^{(k)}_{i}(\mathbf{w}, \mathbf{w}^{(t-1)}),
\end{align}
where
\begin{align}
\Delta \hat{\Phi}^{(k)}_{i}(\mathbf{w}, \mathbf{w}^{(t-1)}) 
:= \hat{\Phi}^{(k)}_{i}(\mathbf{w}) 
- \hat{\Phi}^{(k)}_{i}(\mathbf{w}^{(t-1)}) - \delta.
\end{align}
\end{lemma}

This bound decomposes the drift into a quadratic term capturing the magnitude of forgetting and a linear term weighted by the virtual queues. The first term in the drift upper bound, is non-negative and does not depend on the queue values. Under standard smoothness assumptions on the loss functions, this term can be uniformly bounded by a finite constant, and hence, it does not affect the stability of the virtual queues and can be omitted from the per-round optimization without altering the drift minimization objective up to an additive constant. 
Hence, using the second term of the upper bound derived in the Lemma above, the expression for $\texttt{DPP}^{(t)}_{V,i}$ in \eqref{eq:dpp_1} can be rewritten as
\begin{equation}
    \texttt{DPP}^{(t)}_{V,i} = 
    V \Phi_i(\vecw; \mathcal{D}_i^{(t)})  
    + \sum_{k=1}^{t-1} Q^{(k)}_i[t-1] 
    \Delta \hat{\Phi}^{(k)}_{i}(\vecw, \vecw^{(t-1)}).
    \label{eq:dpp_final}
\end{equation}
The first term encourages learning from the current task via the local loss, scaled by a control parameter $V$ that governs plasticity.\footnote{It is possible to use different $V$ across different clients leading to more flexibility to achieve better performance. However, for the sake of simplicity, we use a fixed $V$ across all clients.} The second term incorporates the accumulated virtual queues, penalizing an increase in replay loss on past tasks, promoting stability. Collectively, this objective adaptively weighs past and present tasks, enabling the model to learn from client data. While the above objective operates locally at each client, the queue-weighted regularization constrains client updates as compared to the previous global update, and this leads  to more consistent and stable global updates upon aggregation. 

\textbf{Algorithm:}~At each task $t$, the optimization proceeds over $C$ communication rounds. At the beginning of round $c$, the server broadcasts the global model $\mathbf{w}^{(t,c-1)}$ to all clients. Each client $i$ initializes its local model as $\mathbf{w}_i^{(t,c,0)} = \mathbf{w}^{(t,c-1)}$ and performs $R$ steps of stochastic gradient descent. The local updates are given by:
\begin{equation}
    \mathbf{w}_i^{(t,c,r+1)} 
    = \mathbf{w}_i^{(t,c,r)} 
    - \eta_t 
    \Bigg(
    V \nabla \Phi_i(\mathbf{w}_i^{(t,c,r)}; \mathcal{B}_i^{(t,c,r)}) 
    + \sum_{k=1}^{t-1} Q^{(k)}_i[t-1] 
    \nabla \hat{\Phi}^{(k)}_{i}(\mathbf{w}_i^{(t,c,r)})
    \Bigg),
\end{equation}
for $r = 0, \dots, R-1$, where $\mathcal{B}_i^{(t,c,r)}$ denotes a mini-batch sampled from $\mathcal{D}_i^{(t)}$. 

A subtle but critical design choice in \textsc{FedQCL} is on \emph{what} the virtual queues measure forgetting relative to, and \emph{when} they are updated. These choices jointly determine whether the queues capture task-level forgetting, client drift, or both. In single-client CL, queues can be updated once per task using the previous model $\mathbf{w}^{(t-1)}$ as reference. In federated settings, model updates are performed iteratively within each task through multiple rounds of local training and aggregation. Hence, we propose to update the virtual queue in every communication round as the underlying learning dynamics evolve continuously across rounds. Updating the queue per round allows for fine-grained tracking of these incremental violations, enabling timely correction through the drift-plus-penalty mechanism. In contrast, updating the queue only at task boundaries would aggregate these effects, leading to delayed feedback and weaker control over the accumulation of forgetting. 

After completing local updates, each client sends $\mathbf{w}_i^{(t,c)} := \mathbf{w}_i^{(t,c,R)}$ to the server. The server aggregates the local models to obtain:
\begin{equation}
    \mathbf{w}^{(t,c)} = \sum_{i=1}^N p^{(t)}_i \mathbf{w}_i^{(t,c)},
\end{equation}
where $p^{(t)}_i = \frac{n_{i,t}}{\sum_{j=1}^N n_{j,t}}$. After $C$ communication rounds, the final global model for task $t$ is given by $\mathbf{w}^{(t)} = \mathbf{w}^{(t,C)}$.

The updated global model $\mathbf{w}^{(t)}$ is then broadcast to all clients and serves as the initialization for the next task. The complete algorithm is provided in Algorithm~\ref{alg:FedQCL} in the supplementary.

\section{Empirical Results}
We evaluate the proposed \textsc{FedQCL} framework in an FCL setting, where data distributions evolve over time, and clients participate across communication rounds. Unlike task-bound formulations, our method operates at the granularity of communication rounds, enabling fine-grained control of forgetting through per-round queue updates. In the sequel, we compare the proposed framework against representative FCL baselines spanning replay, regularization, and gradient projection methods under non-IID and streaming data regimes. We also provide ablation studies with respect to varying levels of heterogeneity, hyperparameters and memory buffer size.

\subsection{Experimental Settings}

We evaluate the proposed method on three widely used image classification benchmarks: CIFAR-10, CIFAR-100~\cite{krizhevsky2009learning}, and TinyImageNet \cite{Le2015TinyIV}. CIFAR-10 consists of 10 classes with 60{,}000 images, while CIFAR-100 contains 100 classes with the same number of images as CIFAR-10. TinyImageNet is a more challenging dataset with 200 classes and higher visual diversity.
 
To simulate the continual learning setting, each dataset is divided into a sequence of disjoint tasks, where each task contains a subset of classes. Specifically, Split-CIFAR-10 is divided into 5 disjoint tasks with 2 classes each, Split-CIFAR-100 into 5 tasks with 20 classes each, and Split-TinyImageNet into 10 tasks with 20 classes each by default. The full set of classes is partitioned into $T$ tasks with an equal number of classes per task, and the model is trained sequentially on these tasks. During training, the model has access only to the data of the current task, while data from previous tasks is available only through a limited memory buffer. Unless specified otherwise, we use a federated setting with $N{=}5$ clients, where data is distributed across clients using a Dirichlet distribution with concentration parameter $\alpha{=}10.0$. Each task is trained for $C{=}10$ communication rounds, with $R{=}5$ local epochs per client per round. A fixed total memory budget of $M{=}500$ samples per client is maintained across
all methods for fair comparison. All experiments are conducted under the {task-incremental} continual
learning setting, where task identity is available during both training and inference, allowing the model to select the appropriate task-specific output head. We use ResNet-18 as the backbone architecture for all experiments due to its widespread adoption in CL literature. The model is trained with SGD using standard data augmentation. Detailed hyperparameters are provided in the supplementary material.

\textbf{Metrics:}~We assess the performance of the proposed algorithm using the server model's test accuracy on a sequence of $T$ tasks, and the forgetting, which are defined below: (a) \textbf{Average Accuracy} (Acc): The average accuracy is computed using $\text{Acc} := \frac{1}{T} \sum_{j=1}^{T} a_{T,j}$, where $a_{t,j}$ is the accuracy of the model obtained after observing task $t$ on task $j <t$. (b) \textbf{Forgetting} (Forget $\in [-1,1]$): We capture the forgetting factor using $\text{Forg} := \frac{1}{T-1} \sum_{j=1}^{T-1} \left[\max_{l \in [T-1]} (a_{l,j} - a_{T,j})\right]$, where the difference $\max_{l \in [T-1]} (a_{l,j} - a_{T,j})$ captures the extent to which previously learned knowledge is overwritten by subsequent learning. 

\subsection{Baselines}

We compare our approach against a diverse set of representative baselines spanning rehearsal-based, regularization-based, and generative replay methods, along with FCL approaches.

\textbf{Centralized CL Methods Extended to Federated Learning:}~
We evaluate federated extensions of standard continual learning approaches, where each client performs local continual updates and a central server aggregates the models using standard FedAvg protocols. \textbf{Elastic Weight Consolidation (EWC)}~\cite{EWC} regularizes training on new tasks by penalizing changes to parameters that are important for previously learned tasks, using a Fisher information–based importance estimate. \textbf{Episodic Replay (ER)}~\cite{ER} maintains a memory buffer of past samples and updates the model using both current data and replayed instances, while \textbf{DER/DER++}~\cite{der} extend this by storing past logits and enforcing consistency during replay, with DER++ additionally incorporating label supervision to further stabilize learning.
\textbf{Learning without Forgetting (LwF)}~\cite{lwf} is a regularization-based approach that uses knowledge distillation to constrain the current model’s predictions to remain close to those of a previously trained model.

\textbf{FCL Methods:}~
We further compare with methods specifically designed for federated continual learning (FCL). These approaches explicitly account for both data heterogeneity and sequential task learning. \textbf{BI}~\cite{Biserra2025federated} is an online FCL method which employs uncertainty-aware memory management to prioritize storing informative samples with high uncertainty for replay. Other representative approaches include \textbf{FOT}~\cite{fot}, which is a federated continual learning framework that mitigates global catastrophic forgetting by restricting local updates to the orthogonal complement of the space spanned by previously learned tasks' gradients. We also compare with \textbf{Class-Conditional Variational Replay (CCVR)}~\cite{ccvrluo2021no} which is a classifier calibration-based method that models each class as a Gaussian distribution in the feature space. It generates synthetic features from these distributions to recalibrate the classifier, enabling effective replay with reduced memory while mitigating the effects of data heterogeneity and forgetting.  Although not explicitly designed for FCL, its data-free replay mechanism makes it applicable to the FCL setting.

\textbf{Standard Federated Learning Baselines:}~
Finally, we include classical federated learning methods without an explicit CL design. \textbf{FedAvg}~\cite{FedAvg} aggregates model parameters from multiple clients via weighted averaging and \textbf{FedProto}~\cite{tan2022fedproto} exchanges class prototypes instead of full model parameters to reduce communication overhead.


\subsection{Comparison with Baselines}
\label{sec:main_comparison}

\begin{table*}[t]
\centering
\caption{Task Incremental Learning Results (Mean $\pm$ Std over 3 seeds) under non-IID data distribution with Dirichlet parameter $\alpha = 10.0$, 5 clients, and a total memory buffer size of 500 samples per client.}
\label{tab:task_incremental}
\renewcommand{\arraystretch}{1.15}
\resizebox{\textwidth}{!}{
\begin{tabular}{l|cc|cc|cc}
\toprule
\multirow{2}{*}{\textbf{Method}} 
& \multicolumn{2}{c|}{\textbf{Split-CIFAR-10}} 
& \multicolumn{2}{c|}{\textbf{Split-CIFAR-100}} 
& \multicolumn{2}{c}{\textbf{Split-TinyImageNet}} \\

& Acc (\%) $\uparrow$ & Forget $\downarrow$ 
& Acc (\%) $\uparrow$ & Forget $\downarrow$ 
& Acc (\%) $\uparrow$ & Forget $\downarrow$ \\
\midrule

FedAvg (2017) & 70.21 $\pm$ 0.75 & 0.12 $\pm$ 0.006 & 29.10 $\pm$ 0.29 & 0.12 $\pm$ 0.001 & 10.09 $\pm$ 0.36 & 0.20 $\pm$ 0.002 \\
FedProto (2022) & 73.03 $\pm$ 0.01 & 0.23 $\pm$ 0.018 & 35.07 $\pm$ 0.01 & 0.27 $\pm$ 0.009 & 19.14 $\pm$ 0.01 & 0.31 $\pm$ 0.01 \\

\addlinespace[4pt]
\midrule

EWC-FL (2017) & 73.51 $\pm$ 0.01 & 0.23 $\pm$ 0.01 & 37.54 $\pm$ 0.01 & 0.24 $\pm$ 0.03 & 21.43 $\pm$ 0.01 & 0.29 $\pm$ 0.003 \\
LwF-FL (2017) & 73.42 $\pm$ 0.01 & 0.23 $\pm$ 0.008 & 36.76 $\pm$ 0.01 & 0.31 $\pm$ 0.01 & 19.38 $\pm$ 0.01 & 0.38 $\pm$ 0.01 \\
ER-FL (2019) & 73.37 $\pm$ 0.02 & 0.23 $\pm$ 0.01 & 37.74 $\pm$ 0.01 & 0.24 $\pm$ 0.007 & 20.99 $\pm$ 0.01 & 0.23 $\pm$ 0.01 \\
DER-FL (2020) & {83.82 $\pm$ 0.52} & {0.07 $\pm$ 0.008} & {48.07 $\pm$ 0.55} & {0.08 $\pm$ 0.008} & {30.62 $\pm$ 1.32} & {0.18 $\pm$ 0.006} \\
DER++-FL (2020) & {83.89 $\pm$ 1.29} & {0.07 $\pm$ 0.01} & \underline{51.04 $\pm$ 0.85} & {0.06 $\pm$ 0.01} & {31.02 $\pm$ 1.01} & {0.17 $\pm$ 0.007} \\

\addlinespace[4pt]
\midrule

CCVR (2021) & 74.78 $\pm$ 0.78 & 0.22 $\pm$ 0.006 & 43.58 $\pm$ 1.58 & 0.16 $\pm$ 0.01 & 21.47 $\pm$ 0.94 & 0.31 $\pm$ 0.007 \\
FOT (2023) & 72.50 $\pm$ 3.97 & \textbf{0.02 $\pm$ 0.02} & 41.24 $\pm$ 1.58 & \textbf{0.01 $\pm$ 0.003} & 28.22 $\pm$ 0.09 & \textbf{0.001 $\pm$ 0.0007} \\
BI (2025) & \underline{{84.46 $\pm$ 0.40}} & {0.10 $\pm$ 0.007} & {49.55 $\pm$ 0.60} &
{{0.11 $\pm$ 0.01}} & \underline{31.05 $\pm$  0.48} &
{{0.19 $\pm$ 0.01}} \\

\addlinespace[4pt]
\midrule

\textsc{FedQCL}(Ours) & \textbf{85.71 $\pm$ 1.12} & \underline{0.05 $\pm$ 0.004} & \textbf{54.52 $\pm$ 0.84} & \underline{0.05 $\pm$ 0.004} & \textbf{33.37 $\pm$ 0.76} & \underline{0.13 $\pm$ 0.005} \\

\bottomrule
\end{tabular}
}
\end{table*}

\begin{figure}[t]
    \centering
    \includegraphics[width=0.490\linewidth]{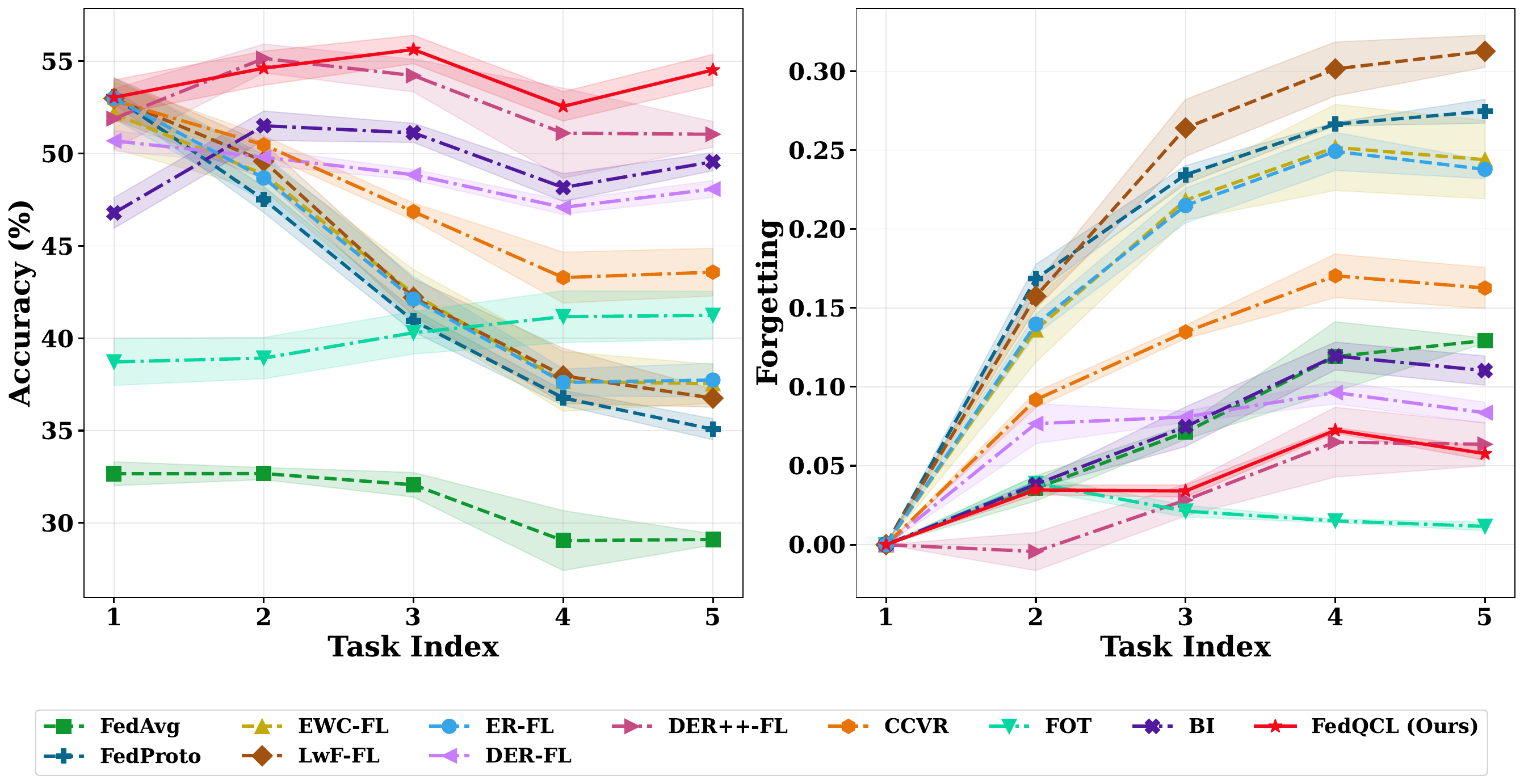}
    \raisebox{4mm}{\includegraphics[width=0.245\linewidth]{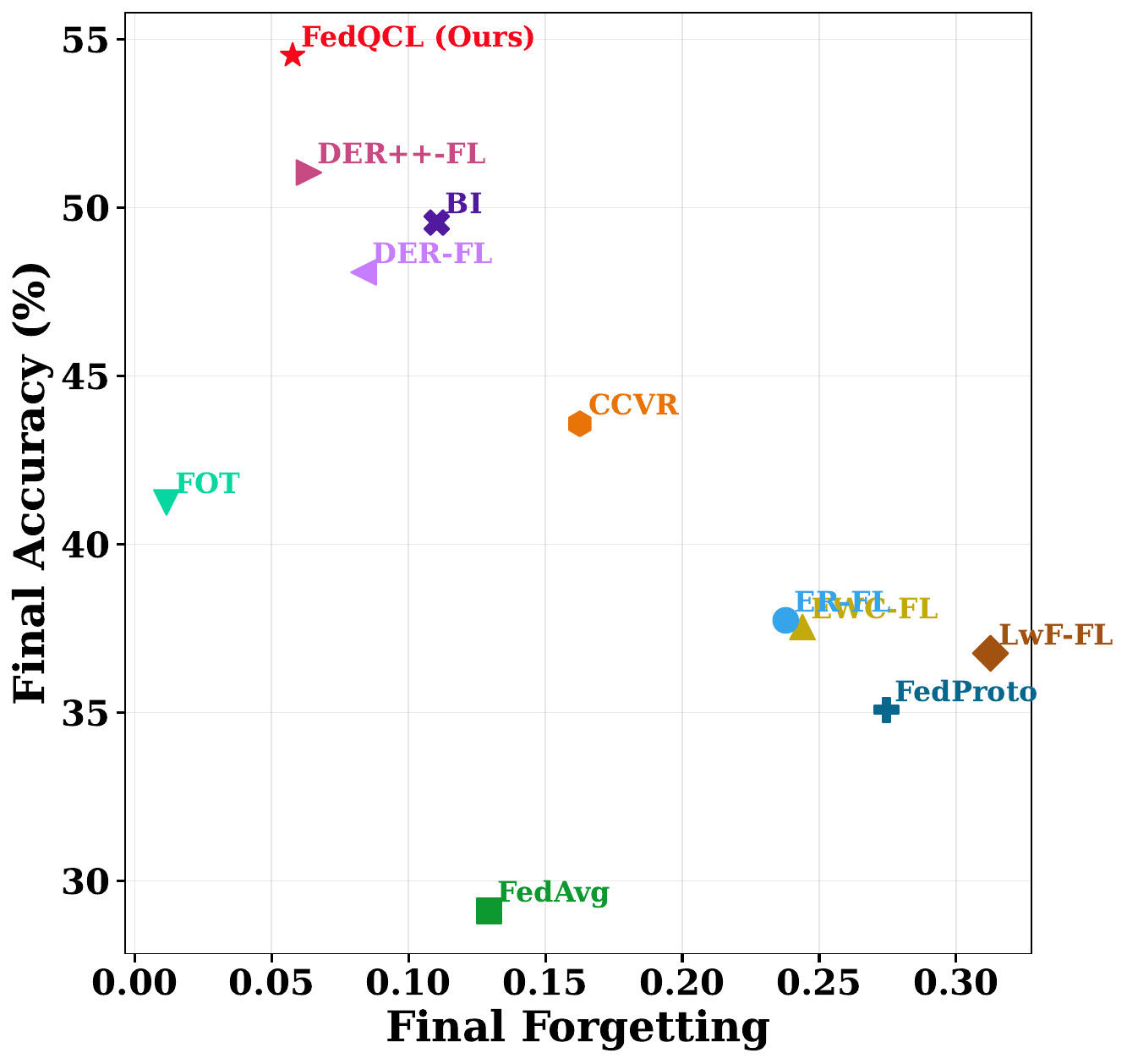}
    \includegraphics[width=0.245\linewidth]{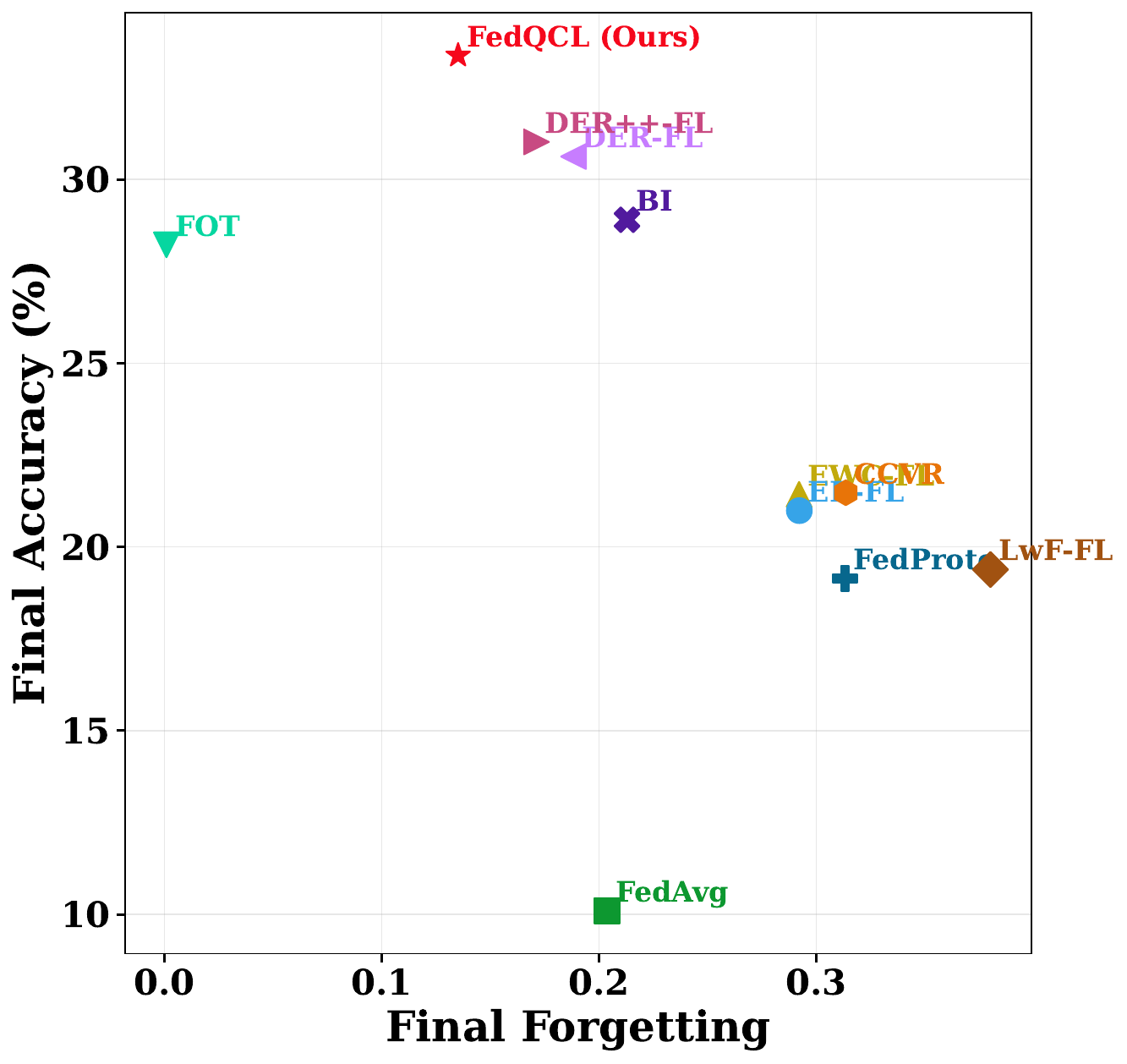}}
    \caption{
        {Left two:} Comparison against baselines on Split-CIFAR-100.
        {Right two:} Plasticity-stability Pareto on Split-CIFAR-100 and
        Split-TinyImageNet.
        Each point represents the final average accuracy vs.\ average forgetting of a method.
    }
    \label{fig:exp1}
\end{figure}

We evaluate \textsc{FedQCL} against all baselines on Split-CIFAR-10, Split-CIFAR-100, and Split-TinyImageNet under the non-IID setting. Results are reported in Table~\ref{tab:task_incremental} and Figure~\ref{fig:exp1}. \textsc{FedQCL} achieves the highest average accuracy across all three benchmark datasets, outperforming the strongest replay baseline DER++-FL. The accuracy curves in Figure~\ref{fig:exp1} (Left) show that \textsc{FedQCL} consistently leads throughout the task sequence, whereas most baselines exhibit a monotonic accuracy decline as the number of observed tasks grows. Regularization-based methods (EWC-FL, LwF-FL) perform on par with FedAvg, confirming that parameter-space constraints alone are insufficient under federated non-IID settings. ER-FL underperforms relative to other replay-based methods across all datasets, likely because it lacks logit-level regularization, making its replay less effective at preventing representation drift under heterogeneous local distributions. \textsc{FedQCL} achieves competitive forgetting scores across all benchmarks. FOT attains near-zero forgetting by projecting gradients orthogonally to past-task subspaces, but at the cost of substantially lower accuracy. 

The Pareto plots in Figure~\ref{fig:exp1} (Right) present final average accuracy against average forgetting for all methods on Split-CIFAR-100 and Split-TinyImageNet. \textsc{FedQCL} occupies the upper-left region of the Pareto front in both datasets, demonstrating the best balance between joint accuracy and forgetting. Methods that achieve low forgetting (FOT) do so by sacrificing plasticity, placing them in the lower portion of the Pareto front. \textsc{FedQCL}'s queue-based penalty dynamically balances these objectives via a single control parameter $V$, avoiding the rigid trade-offs imposed by
projection-based or highly conservative aggregation strategies. Additional results are provided in supplementary~\ref{sec:baselines_comp_supple}.

\begin{figure}[t]
    \centering
    
    \begin{subfigure}{0.49\linewidth}
        \centering
        \includegraphics[width=\linewidth]{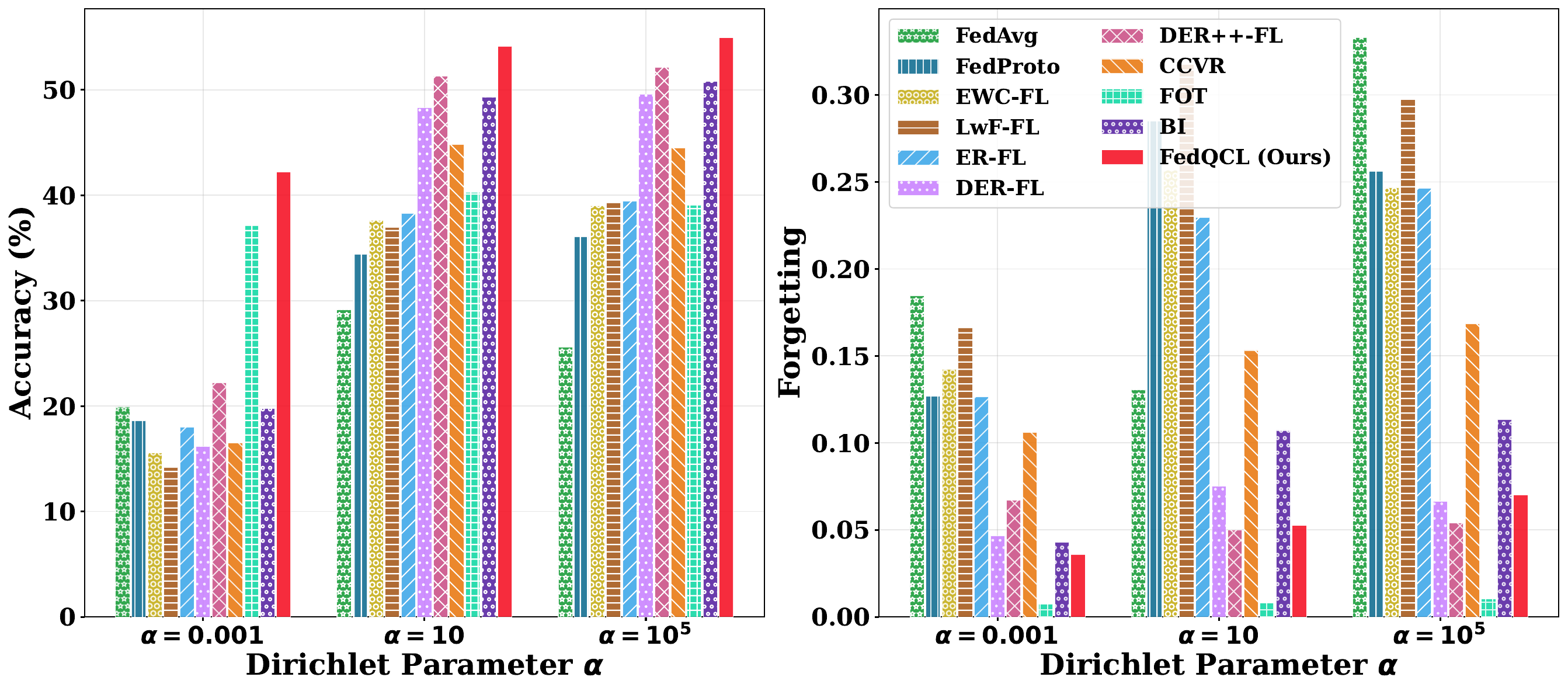}
    \end{subfigure}
    \hfill
    \begin{subfigure}{0.49\linewidth}
        \centering
        \includegraphics[width=\linewidth]{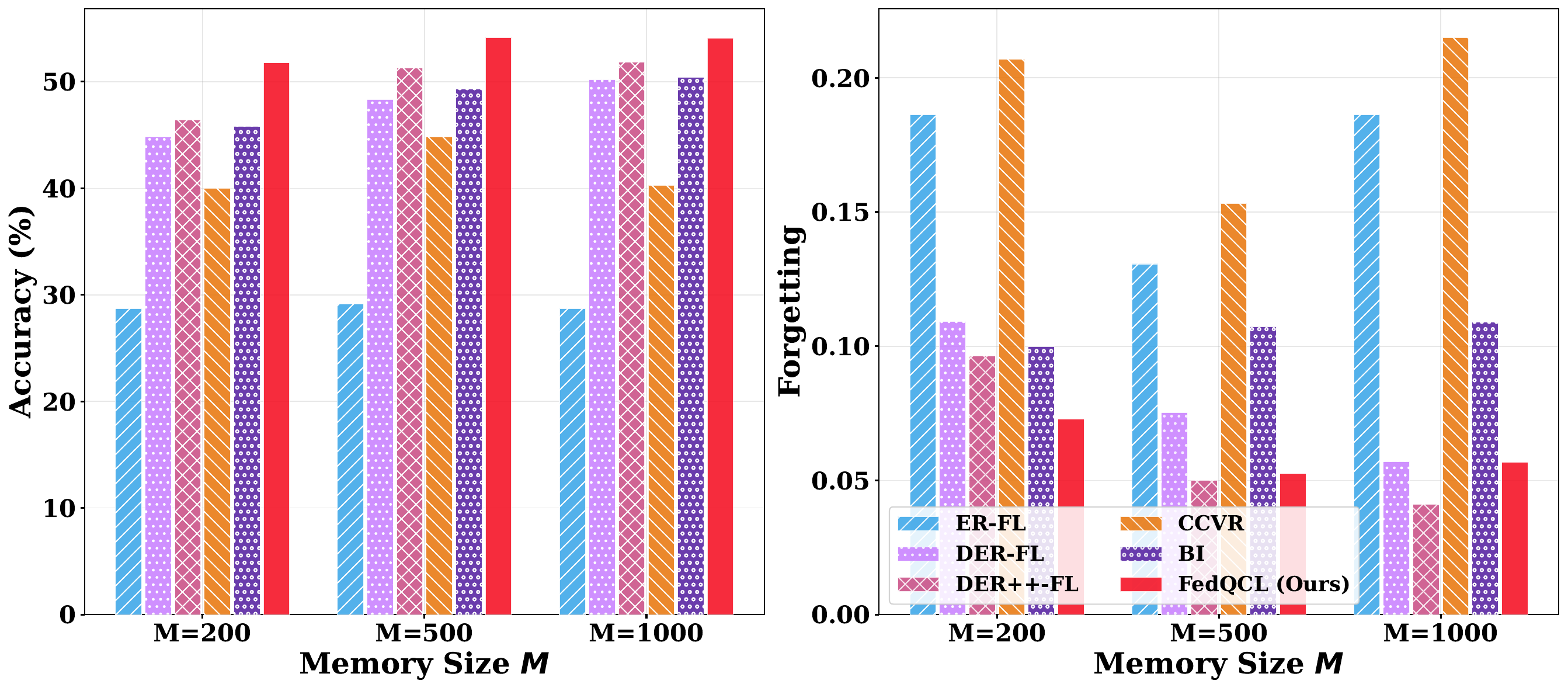}
    \end{subfigure}

    \caption{
        {Ablation on Split-CIFAR-100.}
        Left: Effect of data heterogeneity ($\alpha$), where accuracy (left) and forgetting (right) are shown for $\alpha \in \{0.001,\,10,\,10^5\}$.
        Right: Effect of memory buffer size ($M$), where accuracy (left) and forgetting (right) are reported for $M \in \{200,\,500,\,1000\}$ exemplars per client.
    }
    \label{fig:alpha_memory_combined}
\end{figure}

\subsection{Varying data heterogeneity}
\label{sec:alpha_sweep}


To assess robustness under different levels of distributional shift, we vary the Dirichlet concentration parameter $\alpha\in\{0.001,\,10,\,10^5\}$, spanning highly heterogeneous to near-IID regimes. Results on Split-CIFAR-100 are shown in Figure~\ref{fig:alpha_memory_combined} (Left). \textsc{FedQCL} maintains stable accuracy across all three regimes, with only modest degradation under extreme heterogeneity ($\alpha{=}0.001$). At this setting, most baselines suffer significant accuracy drops because highly imbalanced local data exacerbates client drift, making consistent global updates difficult. \textsc{FedQCL}'s robustness is attributed to the virtual queue mechanism, which anchors forgetting measurement to the shared global reference model $\vecw^{(t-1)}$ throughout all $C$ rounds of task $t$, implicitly discouraging inconsistent cross-client updates without explicit drift-correction overhead. \textsc{FedQCL} maintains low and stable forgetting across all three settings, confirming that the queue-based regularization adapts effectively to varying
degrees of data heterogeneity.

\subsection{Varying Memory Buffer Size}
\label{sec:memory_sweep}







We evaluate sensitivity to the replay buffer size $M\in\{200,\,500,\,1000\}$ per client on Split-CIFAR-100. Results are shown in Figure~\ref{fig:alpha_memory_combined} (Right). \textsc{FedQCL} consistently outperforms baselines at all memory budgets, including the most constrained setting of $M{=}200$. As expected, accuracy improves for all replay-based methods as $M$ increases, since a larger buffer provides better coverage of past-task distributions. Notably, the relative performance gap between \textsc{FedQCL} and DER++-FL remains roughly stable across memory sizes, indicating that \textsc{FedQCL}'s advantage stems primarily from its principled queue-based regularization rather than solely from improved replay quality.


\subsection{Ablation Studies}

\subsubsection{Effect of the hyperparameter $V$}
\label{sec:ablation_V}
 
\begin{figure}[t]
    \centering
    
    \begin{subfigure}{0.24\linewidth}
        \centering
        \includegraphics[width=\linewidth]{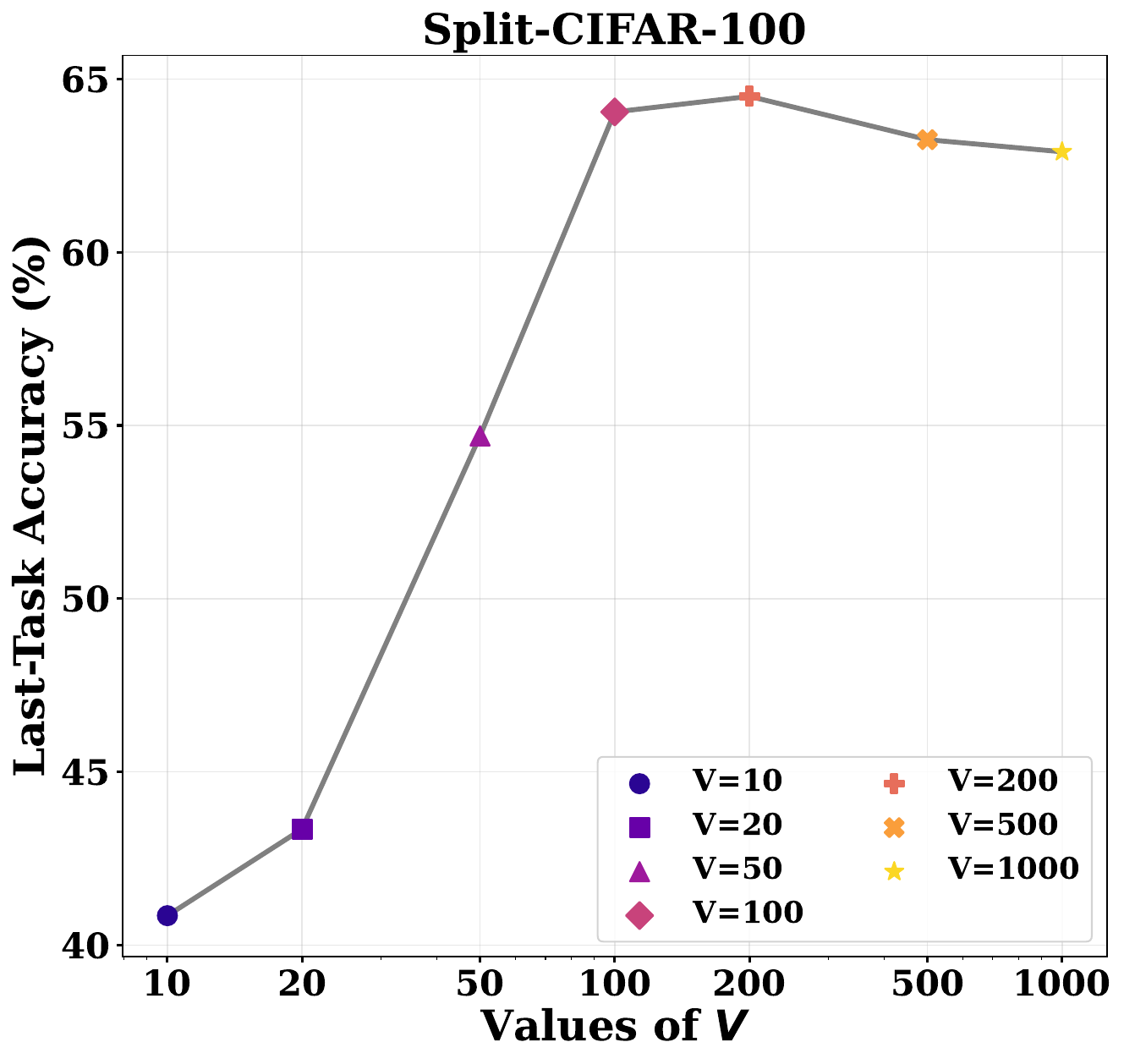}
    \end{subfigure}
    \hfill
    \begin{subfigure}{0.24\linewidth}
        \centering
        \includegraphics[width=\linewidth]{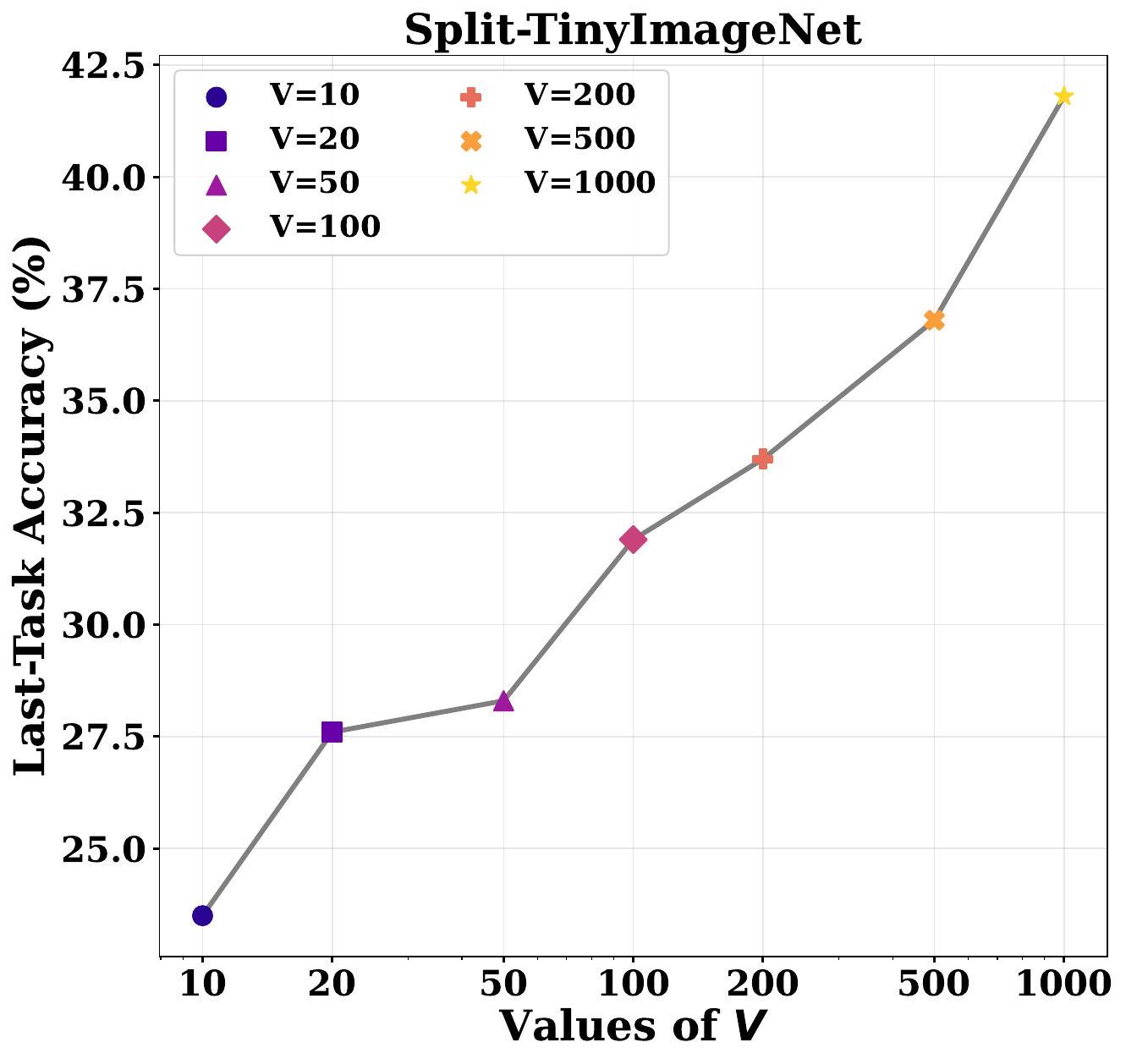}
    \end{subfigure}
    \hfill
    \begin{subfigure}{0.24\linewidth}
        \centering
        \includegraphics[width=\linewidth]{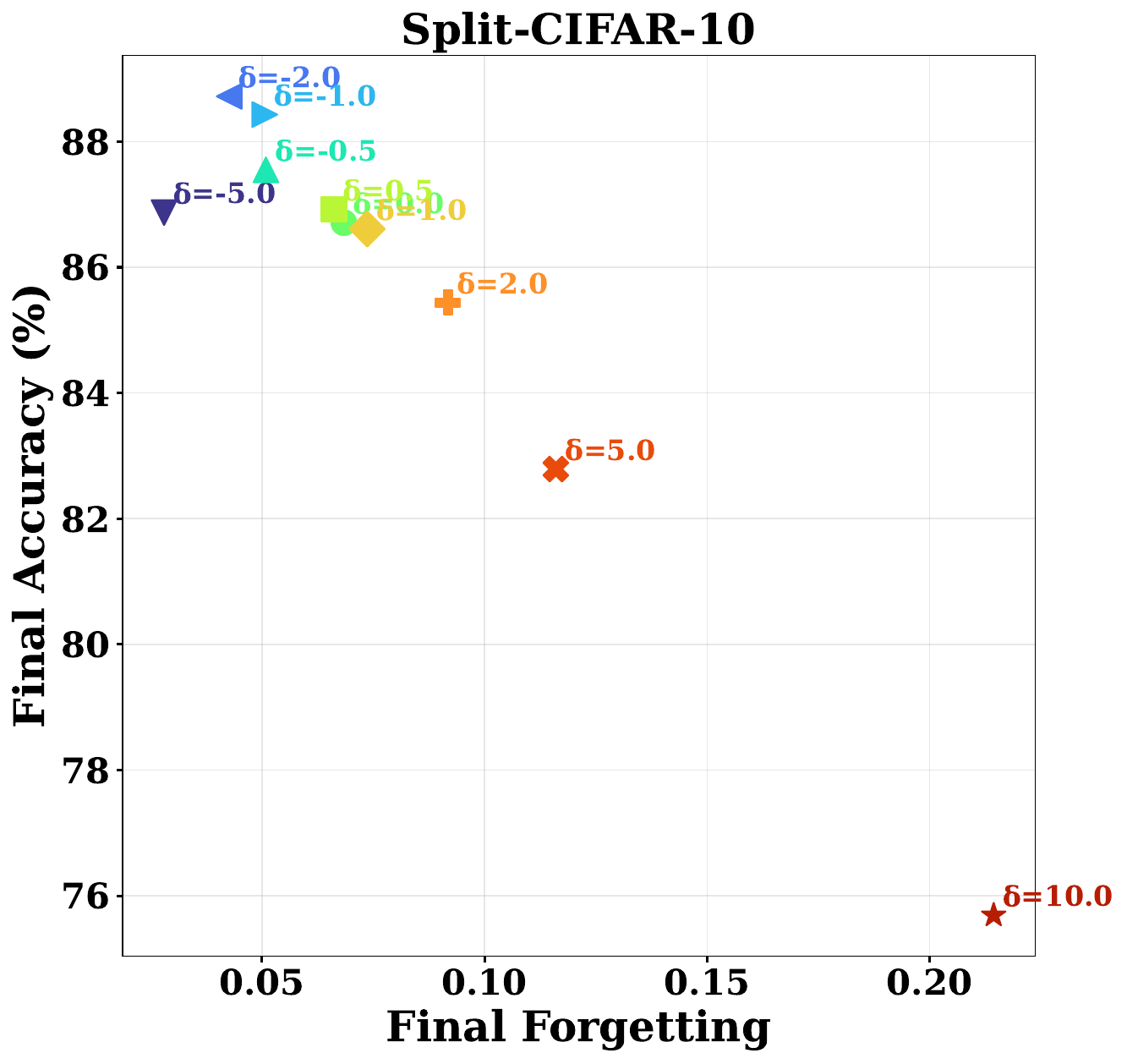}
    \end{subfigure}
    \hfill
    \begin{subfigure}{0.24\linewidth}
        \centering
        \includegraphics[width=\linewidth]{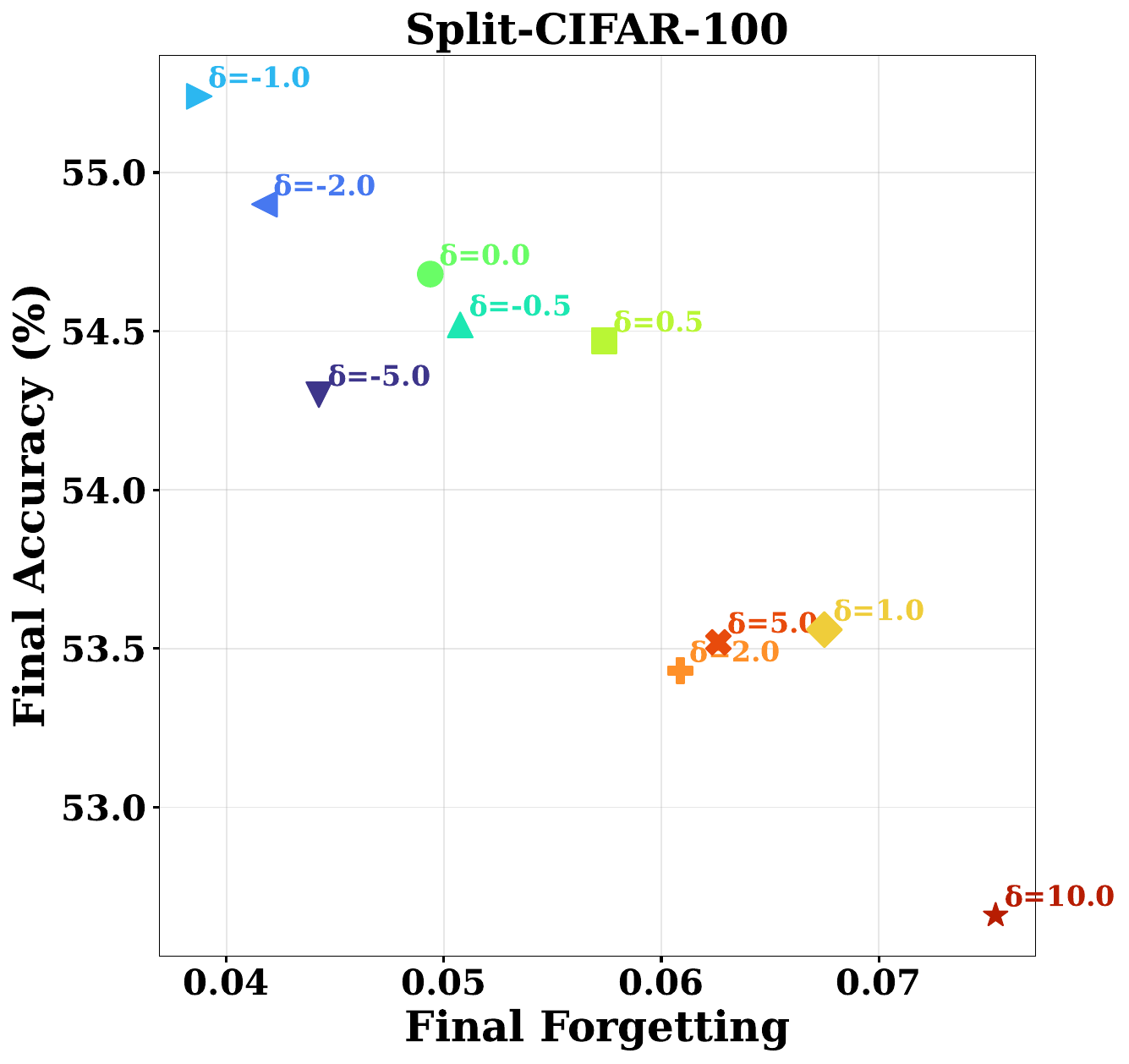}
    \end{subfigure}

    \caption{
        {Hyperparameter analysis of \textsc{FedQCL}.}
        Left: Effect of $V$ on Split-CIFAR-100 and Split-TinyImageNet (accuracy and forgetting vs.\ $V$).
        Right: Effect of $\delta$ shown via plasticity–stability Pareto trade-off on Split-CIFAR-10 and Split-CIFAR-100.
    }
    \label{fig:main_hyperparams}
\end{figure}

{The parameter $V$ controls the balance between plasticity (current-task learning) and stability (forgetting regularization). We vary $V\in\{10,\,20,\,50,\,100,\,200,\,500,\,1000\}$ on Split-CIFAR-100 and Split-TinyImageNet, with Last-Task-Accuracy (LTA) results in Figures~\ref{fig:main_hyperparams} (Left). For very small $V$ (e.g., $V{=}10$), the queue penalty dominates the optimization objective, excessively suppressing adaptation to new tasks and leading to low LTA. As $V$ increases, LTA rises and peaks around $V=100$ to $V=200$ on Split-CIFAR-100, confirming that plasticity improves as the current-task loss gains relative weight. On Split-TinyImageNet, LTA increases monotonically across the full range, as the higher task difficulty requires a larger $V$ to sufficiently relieve the queue penalty. LTA isolates the plasticity effect of $V$ directly. This validates the theoretical design of \textsc{FedQCL}: $V$ acts as a single, interpretable control knob that smoothly interpolates between forgetting-averse and plasticity-focused regimes.}

\begin{figure}[b]
    \centering
    \includegraphics[width=0.80\linewidth]{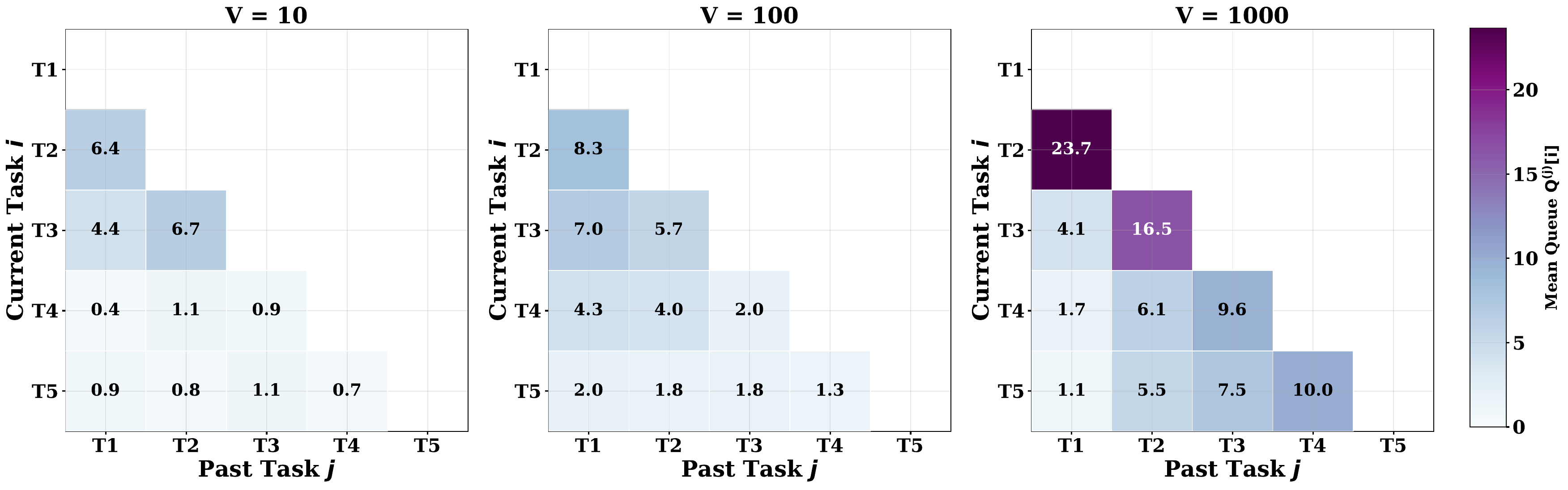}
    \caption{
        Virtual queue matrix $Q^{(j)}[i]$ under varying penalty weight $V$ on Split-CIFAR-100.
        Cell $(i,j)$ represents the queue value for past task $j$
        accumulated during training on task $i$, averaged across clients.
    }
    \label{fig:queue_heatmap_V}
\end{figure}

\subsubsection{Effect of the Hyperparameter $\delta$}
\label{sec:ablation_delta}
 
The parameter $\delta$ defines the allowable per-round increase in replay loss before queues begin to accumulate. We vary $\delta\in\{-5.0,\,-2.0,\,-1.0,\,-0.5,\,0.0,\,0.5,\,1.0,\,2.0,\,5.0,\,10.0\}$ on Split-CIFAR-10 and Split-CIFAR-100, with results in Figures~\ref{fig:main_hyperparams} (Right). Negative $\delta$ imposes stricter constraints by requiring the replay loss to actively decrease relative to the reference model, leading to larger queue values and stronger regularization pressure. As $\delta$ increases toward positive values, the constraint becomes more lenient: forgetting increases while accuracy can benefit marginally from the additional plasticity. The Pareto plots reveal an optimal region around $\delta\in[-1.0,\,1.0]$, where both accuracy and forgetting are well-balanced. A sufficiently large positive $\delta$ renders the queue constraint inactive, effectively reducing the method to standard replay-based training. In contrast, highly negative values of $\delta$ impose overly restrictive constraints, limiting plasticity and consequently degrading predictive performance. These results confirm that $\delta$ provides an interpretable secondary control over the stability-plasticity trade-off, complementing $V$.

\subsubsection{Queue Dynamics}
\label{sec:queue_dynamics}

Figure~\ref{fig:queue_heatmap_V} shows the virtual queue matrix $Q^{(j)}[i]$ under varying $V$ on Split-CIFAR-100. Large $V$ (e.g., $V{=}1000$) results in elevated queue values concentrated on early tasks, since the dominance of the current-task loss leaves less capacity for the queue penalty to suppress forgetting. At moderate $V$ (e.g., $V{=}100$), the queue values are more uniformly distributed and smaller in magnitude, indicating that the algorithm more evenly manages forgetting pressure across all past tasks. Analysis of $\delta$ and $\alpha$ is provided in the supplementary.

\section{Conclusions}
We proposed \textsc{FedQCL}, a control-theoretic framework that explicitly regulates forgetting in FCL. By formulating FCL as a stochastic control problem, we introduced virtual queues to track and constrain the accumulation of forgetting across clients and communication rounds, allowing a DPP-based algorithm. A key insight is that forgetting evolves continuously at the level of communication rounds in FCL, motivating per-round queue updates. Maintaining queues locally at each client enabled effective regulation without additional communication overhead while being privacy-preserving, making the approach scalable to realistic federated settings. Experimental results demonstrated that \textsc{FedQCL} achieves competitive accuracy while significantly reducing forgetting under distributed, heterogeneous, and non-stationary data distributions. Overall, this work highlights the importance of treating FCL as a dynamic control problem, enabling principled and efficient regulation of the stability-plasticity trade-off.

\section*{Acknowledgments}
This work was supported in part by LightMetrics Pvt. Ltd., India, and by the Ministry of Electronics and Information Technology (MeitY), Government of India, through the IndiaAI Mission.

\bibliographystyle{tmlr}
\bibliography{refs.bib}
\newpage

\section{Supplementary section}
\subsection{Additional Theoretical Details}

\subsubsection{Proof of Lemma~1}

From \eqref{eq:queue_update}, we know that the virtual queues evolve as:
\begin{align}
Q^{(k)}_{i}[t] = \max \left\{ Q^{(k)}_{i}[t-1] + \Delta \hat{\Phi}^{(k)}_{i}(\mathbf{w}, \mathbf{w}^{(t-1)}), \, 0 \right\}.
\end{align}

Using $(\max\{x,0\})^2 \leq x^2$, we obtain:
\begin{align}
\left(Q^{(k)}_{i}[t]\right)^2 
&\leq \left( Q^{(k)}_{i}[t-1] + \Delta \hat{\Phi}^{(k)}_{i}(\mathbf{w}, \mathbf{w}^{(t-1)}) \right)^2 \\
&= \left(Q^{(k)}_{i}[t-1]\right)^2 
+ \left( \Delta \hat{\Phi}^{(k)}_{i}(\mathbf{w}, \mathbf{w}^{(t-1)}) \right)^2 
+ 2 Q^{(k)}_{i}[t-1] \, \Delta \hat{\Phi}^{(k)}_{i}(\mathbf{w}, \mathbf{w}^{(t-1)}).
\end{align}

Summing over $k$ and multiplying by $\frac{1}{2}$:
\begin{align}
L_i[t] 
&\leq L_i[t-1] 
+ \frac{1}{2} \sum_{k=1}^{t-1} \left( \Delta \hat{\Phi}^{(k)}_{i}(\mathbf{w}, \mathbf{w}^{(t-1)}) \right)^2
+ \sum_{k=1}^{t-1} Q^{(k)}_{i}[t-1] \, \Delta \hat{\Phi}^{(k)}_{i}(\mathbf{w}, \mathbf{w}^{(t-1)}).
\end{align}

Rearranging gives the desired bound.
\subsubsection{Algorithm Table}

\begin{algorithm}[h!]
\caption{\textbf{Federated Queue regulated Continual Learning (\textsc{FedQCL})}} 
\label{alg:FedQCL}
\begin{algorithmic}[1]
    \State \textbf{Initialize:} $Q^{(k)}_i[0] = 0, \ \forall i,k$, initialize global model $\mathbf{w}^{(0)}$, $\delta \geq 0$, $V > 0$.

\For{$t = 1, 2, \ldots, T$} \Comment{Tasks}

    \For{$c = 1, 2, \ldots, C$} \Comment{Communication Rounds}
    
        \State \textbf{Server:} Broadcast $\mathbf{w}^{(t,c-1)}$ to all clients.
        
        \For{each client $i \in [N]$ \textbf{in parallel}}
        
            \State Initialize local model: $\mathbf{w}_i^{(t,c,0)} = \mathbf{w}^{(t,c-1)}$.
            
            \For{$r = 0, 1, \ldots, R-1$} \Comment{Local Steps}
            
                \State Sample mini-batch $\mathcal{B}_i^{(t,c,r)}$ from $\mathcal{D}_i^{(t)}$.
                
                \State \textbf{Local DPP SGD update:}
                \begin{align*}
                \mathbf{w}_i^{(t,c,r+1)} 
                &= \mathbf{w}_i^{(t,c,r)} 
                - \eta_t 
                \Bigg( 
                V \nabla \Phi_i(\mathbf{w}_i^{(t,c,r)}; \mathcal{B}_i^{(t,c,r)})  + \sum_{k=1}^{t-1} Q^{(k)}_i[t-1] 
                \nabla \hat{\Phi}^{(k)}_{i}(\mathbf{w}_i^{(t,c,r)}) 
                \Bigg).
                \end{align*}
                
            \EndFor
            
            \State Set $\mathbf{w}_i^{(t,c)} = \mathbf{w}_i^{(t,c,R)}$.
            
            \State \textbf{Queue update (client-side):}
            \begin{equation*}
            Q^{(k)}_i[t] = 
            \max \left\{ 
            Q^{(k)}_i[t-1] + 
            \hat{\Phi}^{(k)}_{i}(\mathbf{w}_i^{(t,c)}) - \hat{\Phi}^{(k)}_{i}(\mathbf{w}^{(t-1)}) - \delta, \, 0 
            \right\}, 
            \ \forall k < t.
            \end{equation*}
            
        \EndFor
        
        \State \textbf{Server aggregation:}
        \begin{equation*}
        \mathbf{w}^{(t,c)} = \sum_{i=1}^N p^{(t)}_i \mathbf{w}_i^{(t,c)}.
        \end{equation*}
        
    \EndFor
\EndFor

$\mathbf{w}^{(t)} = \mathbf{w}^{(t,C)}$
\end{algorithmic}
\end{algorithm}

\subsubsection{Complexity Analysis}

\begin{table*}[t]
\centering
\caption{%
  Complexity comparison per communication round per client.
  $N$: total clients; $C$: communication rounds per task; $T$: tasks;
  $M$: replay buffer size; $d$: model parameter dimension; $B$: mini-batch size; $B_m$: memory replay batch size;
  $K$: number of past tasks (grows with $T$); $P$: number of augmentation passes.
}
\label{tab:complexity}
\resizebox{\textwidth}{!}{%
\begin{tabular}{l l l l l}
\toprule
\textbf{Method} & \textbf{Client Storage} & \textbf{Client Computation} & \textbf{Communication (per round)} & \textbf{Extra Cost at Task Boundary} \\
\midrule

\textsc{FedQCL} (Ours)
  & $\mathcal{O}(M + K + d)$ per client
  & $\mathcal{O}((B + K \cdot B_m) \cdot d)$ per round
  & $\mathcal{O}(d)$
  & $\mathcal{O}(C)$ queue updates per task\\
  & \footnotesize{(buffer $+$ prev.\ task model $+$ $K$ scalar queues)}
  & \footnotesize{(current-task grad $+$ $K$ replay grads)}
  & \footnotesize{(same as {FedAvg})}
  & \footnotesize{queues updated in-place)} \\[4pt]

FOT
  & $\mathcal{O}(0)$ client-side
  & $\mathcal{O}(B \cdot d)$
  & $\mathcal{O}(d)$ per round
  & $\mathcal{O}(d \cdot s_{\max})$ upload $+$ SVD $\mathcal{O}(d \cdot s_{\max}^2)$ at server \\
  & \footnotesize{(server stores orthogonal subspace $\mathcal{O}(\sum_\ell d^\ell \cdot r^\ell)$)}
  & \footnotesize{(standard SGD; projection is server-side)}
  & \footnotesize{({FedAvg}; $+\,\mathcal{O}(d \cdot s_{\max})$ at task boundary)}
  & \footnotesize{(one extra round per task; SVD cost at server.)} \\[4pt]

BI
  & $\mathcal{O}(M)$ per client
  & $\mathcal{O}(P \cdot B \cdot d)$ per round
  & $\mathcal{O}(d)$
  & $\mathcal{O}(0)$ \\
  & \footnotesize{(class-balanced buffer)}
  & \footnotesize{($P$ forward passes per round $+$ replay grad)}
  & \footnotesize{(same as {FedAvg})}
  & \footnotesize{(no extra round at task boundary)} \\

DER-FL
  & $\mathcal{O}(M \cdot (1 + C_{out}))$
  & $\mathcal{O}((B + B_m) \cdot d)$
  & $\mathcal{O}(d)$
  & $\mathcal{O}(0)$ \\
  & \footnotesize{(buffer samples $+$ stored logits per sample)}
  & \footnotesize{(SGD $+$ logit consistency on replayed samples)}
  & \footnotesize{(FedAvg)}
  & \\[4pt]

DER++-FL
  & $\mathcal{O}(M \cdot (1 + C_{out}))$
  & $\mathcal{O}((B + B_m) \cdot d)$
  & $\mathcal{O}(d)$
  & $\mathcal{O}(0)$ \\
  & \footnotesize{(buffer samples $+$ stored logits per sample)}
  & \footnotesize{(SGD $+$ logit consistency $+$ CE on replayed samples)}
  & \footnotesize{(FedAvg)}
  & \\[4pt]

ER-FL
  & $\mathcal{O}(M)$
  & $\mathcal{O}((B + B_m) \cdot d)$
  & $\mathcal{O}(d)$
  & $\mathcal{O}(0)$ \\
  & \footnotesize{(sample buffer)}
  & \footnotesize{(SGD on current $+$ replayed samples)}
  & \footnotesize{(FedAvg)}
  & \\[4pt]

LwF-FL
  & $\mathcal{O}(d)$
  & $\mathcal{O}(B \cdot d)$
  & $\mathcal{O}(d)$
  & $\mathcal{O}(d)$ \\
  & \footnotesize{(prev.\ task model $w^{(t-1)}$; no buffer)}
  & \footnotesize{(SGD $+$ distillation forward pass on current batch)}
  & \footnotesize{(FedAvg)}
  & \footnotesize{(save model checkpoint at task end)} \\[4pt]

EWC-FL
  & $\mathcal{O}(d)$
  & $\mathcal{O}(B \cdot d)$
  & $\mathcal{O}(d)$
  & $\mathcal{O}(B \cdot d)$ \\
  & \footnotesize{(Fisher diagonal $+$ prev.\ model; no buffer)}
  & \footnotesize{(SGD $+$ EWC penalty; no extra forward pass)}
  & \footnotesize{(FedAvg)}
  & \footnotesize{(Fisher diagonal computation over local data)} \\[4pt]

\midrule

\bottomrule
\end{tabular}%
}
\end{table*}

{We provide a complexity comparison between \textsc{FedQCL} and baselines in Table~\ref{tab:complexity}. The replay-based federated extensions (ER-FL, DER-FL, DER++-FL) and regularization-based methods (LwF-FL, EWC-FL) all follow standard \textsc{FedAvg} aggregation with no additional communication overhead, differing only in their client-side storage and per-round computation. FOT eliminates client storage entirely but introduces a dedicated subspace extraction round at each task boundary, during which clients transmit activation sketches and the server performs SVD, making it the most expensive method at task boundaries. BI avoids task-boundary costs but incurs roughly $2.2\times$ the per-round computation of plain ER due to $P$-fold test-time augmentation~\cite{Biserra2025federated}. \textsc{FedQCL} stores a replay buffer of size $M$ and the previous global model $\mathbf{w}^{(t-1)} \in \mathbb{R}^d$, the latter serving as the reference point for forgetting measurement. Importantly, \textsc{FedQCL} introduces no additional communication rounds, no server-side computation beyond standard aggregation, and no task-boundary overhead. Queue updates are performed in-place at the client within the existing round structure, making \textsc{FedQCL} as communication-efficient as FedAvg.}

\subsection{Additional Experimental Details}

\subsubsection{Implementation Details and Codebases}
\label{sec:implementation}

All experiments are implemented in PyTorch. We use ResNet-18 as the backbone for all methods to ensure a fair comparison. We note that several baselines were originally proposed for class-incremental settings and have been adapted to the task-incremental setting used in this work. For all baselines, we use their recommended hyperparameters where available. The following codebases were used to reproduce the baselines:

\begin{itemize}
    \item {EWC-FL, LwF-FL, DER-FL, DER++-FL, CCVR, FedAvg, FedProto:} 
    Adapted from Fed-Mammoth repository~\cite{salami2024closed_mammoth}
    \item {FOT:} Official implementation available at~\cite{fot}
    \item {BI:} Official implementation available at~\cite{Biserra2025federated}
\end{itemize}

Table~\ref{tab:fedqcl_hyperparams} lists the hyperparameters used in all \textsc{FedQCL} experiments. We adopt a balanced reservoir sampling strategy to maintain a fixed-size replay buffer of $M$ samples per client across the task sequence. The values of hyperparameters $V$ and $\delta$ are selected based on the sensitivity analysis presented in Section~\ref{sec:ablation_V} and~\ref{sec:ablation_delta}.

\begin{table}[h!]
\centering
\caption{Hyperparameters used in \textsc{FedQCL} experiments.}
\label{tab:fedqcl_hyperparams}
\begin{tabular}{lccc}
\toprule
\textbf{Hyperparameter} & \textbf{Split-CIFAR10} & \textbf{Split-CIFAR100} & \textbf{Split-TinyImageNet} \\
\midrule

Learning rate        & 0.005 & 0.005 & 0.005 \\
Optimizer            & SGD  & SGD   & SGD  \\
Local epochs         & 5      & 5      & 5      \\

\midrule
\# Clients           & 5      & 5      & 5      \\
Communication rounds per task  & 10     & 10     & 10     \\

\midrule
\# Tasks             & 5      & 5     & 10     \\
$V$                  & 20      & 200     & 220    \\
Queue initialization & 0      & 0      & 0      \\
$\delta$-parameter   & 1    & 1      & 1      \\

\midrule
Task batch size      & 13     & 13     & 13     \\
Memory batch size      & 13     & 13     & 13     \\

\bottomrule
\end{tabular}
\end{table}

\begin{figure}[t]
    \centering
    \includegraphics[width=0.490\linewidth]{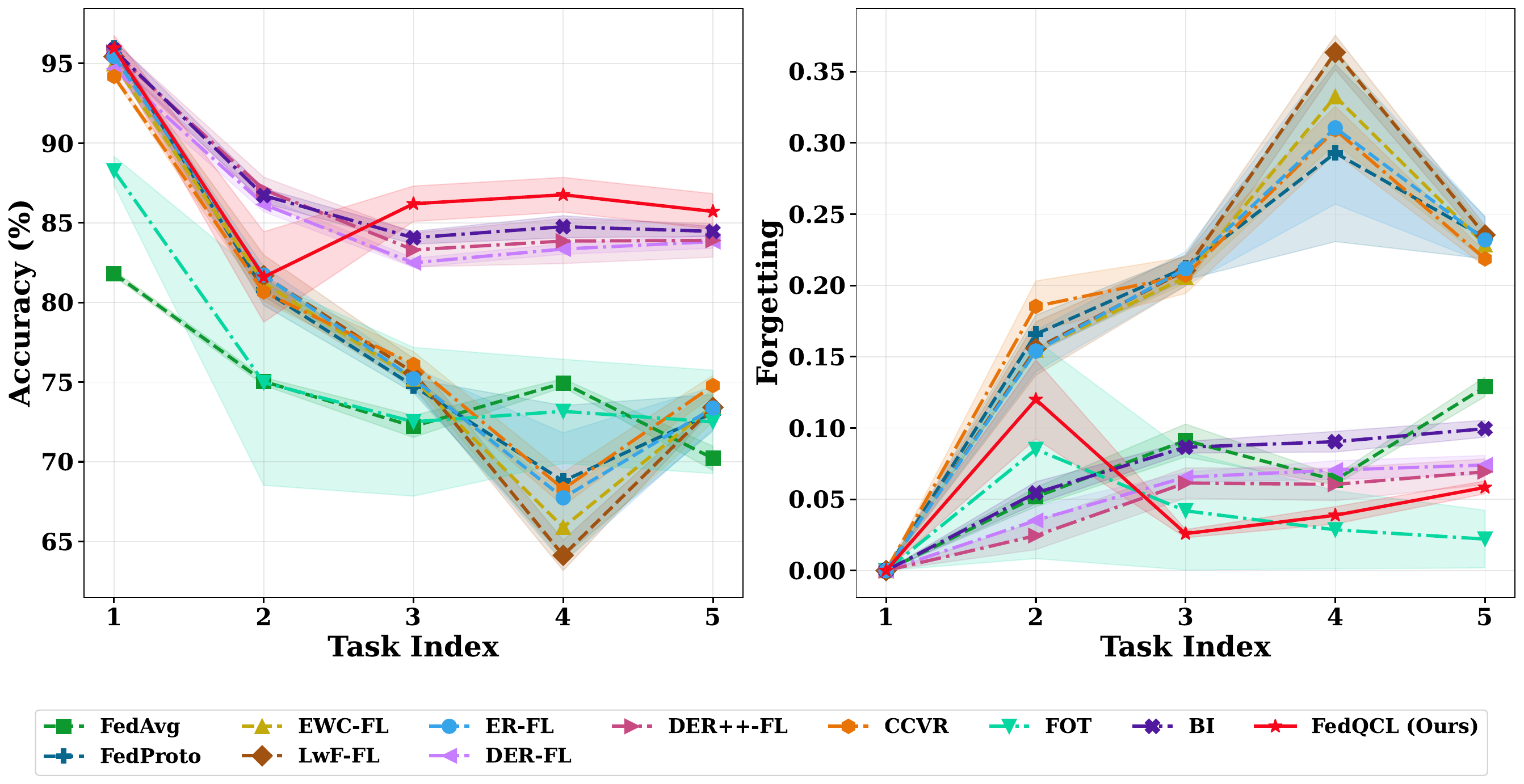}
    \includegraphics[width=0.490\linewidth]{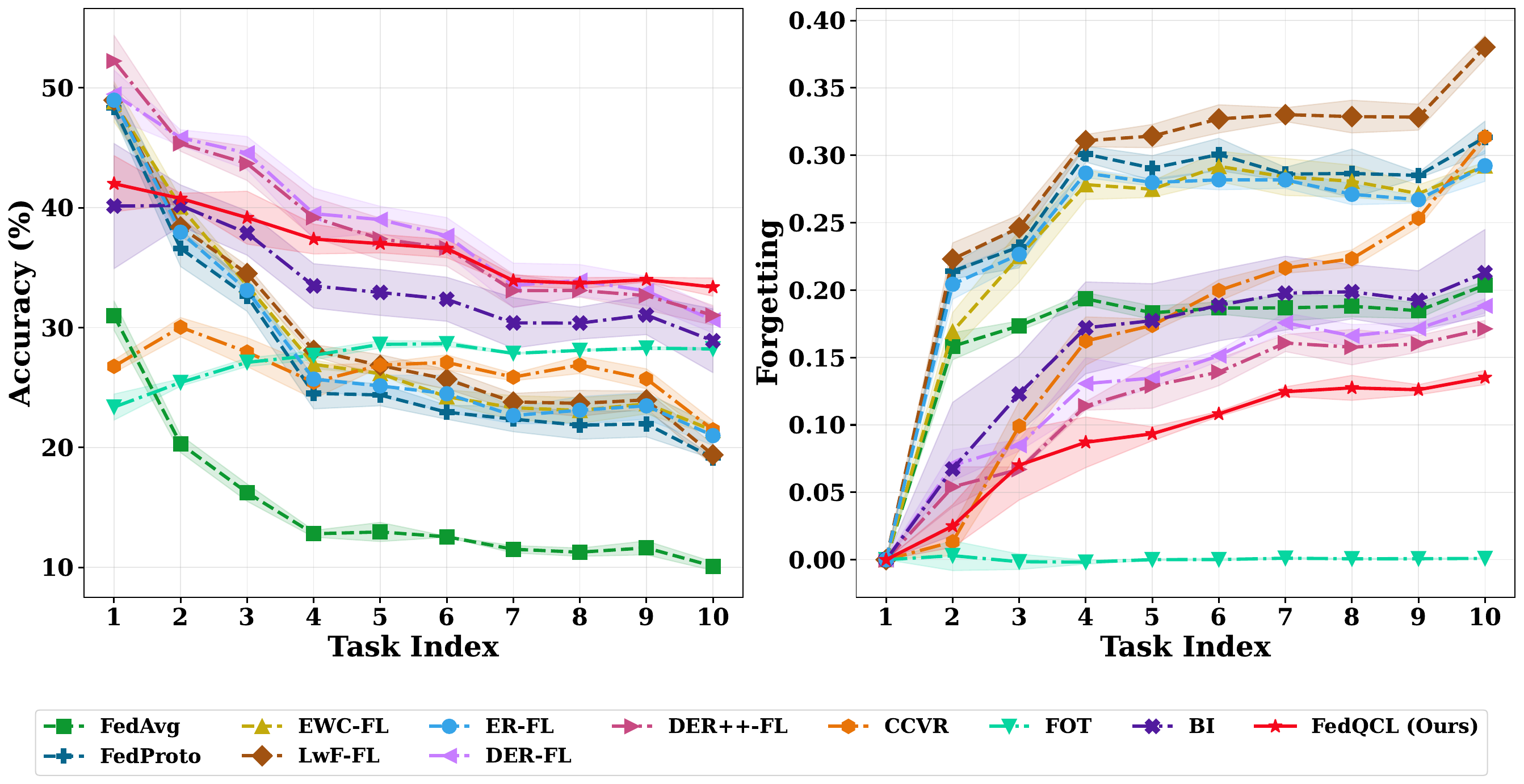}
    
    \caption{
        {Left two:} Comparison against baselines on Split-CIFAR-10.
        {Right two:} Comparison against baselines on Split-TinyImageNet.
    }
    \label{fig:exp1_supple}
\end{figure}
\subsubsection{Comparison with baselines}
\label{sec:baselines_comp_supple}

Figure~\ref{fig:exp1_supple} shows the accuracy and forgetting curves over the task sequence for Split-CIFAR-10 and Split-TinyImageNet, complementing the Split-CIFAR-100 results presented in the main paper (Figure~\ref{fig:exp1}). \textsc{FedQCL} consistently maintains the highest accuracy throughout the task sequence on both datasets, while most baselines exhibit progressive accuracy degradation. Forgetting trends are consistent with the main paper findings.

\begin{table}[h]
\centering
\caption{Class-incremental learning results on Split-CIFAR-100 under near-IID ($\alpha=10^5$) and non-IID ($\alpha=0.1$) client distributions.}
\label{tab:cil_results}
\begin{tabular}{lcccc}
\toprule
\multirow{2}{*}{Method} & \multicolumn{2}{c}{$\alpha = 10^5$} & \multicolumn{2}{c}{$\alpha = 0.1$} \\
\cmidrule(lr){2-3} \cmidrule(lr){4-5}
& Acc (\%) $\uparrow$ & Forget $\downarrow$ & Acc (\%) $\uparrow$ & Forget $\downarrow$ \\
\midrule
FedAvg      & 12.99 & 0.39 & 8.60  & 0.24 \\
EWC         & 16.79 & 0.35 & 9.53  & 0.25 \\
LwF         & 15.75 & 0.49 & 6.48  & 0.26 \\
ER          & 13.93 & 0.37 & 11.69 & 0.26 \\
DER         & 16.20 & 0.43 & 7.60  & \textbf{0.03} \\
DER++       & 24.32 & 0.13 & 14.02 & {0.06} \\
\textbf{\textsc{FedQCL} (Ours)} & \textbf{24.92} & \textbf{0.11} & \textbf{14.66} & 0.24 \\
\bottomrule
\end{tabular}
\end{table}
\subsubsection{Class-Incremental Learning Setting}
\label{sec:cil}

While the main results consider the task-incremental setting with a task oracle at inference, we additionally evaluate \textsc{FedQCL} under the more challenging class-incremental learning (CIL) setting, where task identity is unavailable at test time and the model must discriminate among all classes seen so far. The \textsc{FedQCL} queue mechanism operates directly on replay losses and is agnostic to the availability of task identity, so the drift-plus-penalty regularization objective in equation~\ref{eq:dpp_final} is unchanged across both settings. The only modification required for CIL is at the output level: since a single shared classification head is used, we augment the local objective with a logit-matching term on replayed samples, analogous to~\cite{der}, to enforce output-level consistency across tasks. This term complements the queue-based penalty, which regulates forgetting at the level of task performance rather than individual logits. We evaluate on Split-CIFAR-100 under both a near-IID ($\alpha=10^5$) and a heterogeneous non-IID ($\alpha=0.1$) client data distribution. Results are reported in Table~\ref{tab:cil_results}.

\textsc{FedQCL} outperforms all baselines in average accuracy under both the near-IID and non-IID regimes, confirming that the queue-based regularization remains effective even without task identity at inference. Forgetting is competitive with the strongest baselines in the near-IID setting, though under high heterogeneity ($\alpha=0.1$) the forgetting-accuracy trade-off shifts, with DER and DER++ achieving lower forgetting at the cost of substantially lower accuracy. These results indicate that \textsc{FedQCL}'s queue mechanism generalizes beyond the task-incremental setting and remains a viable regularizer when combined with a shared-head, logit-matching objective.

\subsubsection{Varying Number of Clients}
\label{sec:client_sweep}

\begin{figure}[t]
    \centering
    
    \begin{subfigure}{0.49\linewidth}
        \centering
        \includegraphics[width=\linewidth]{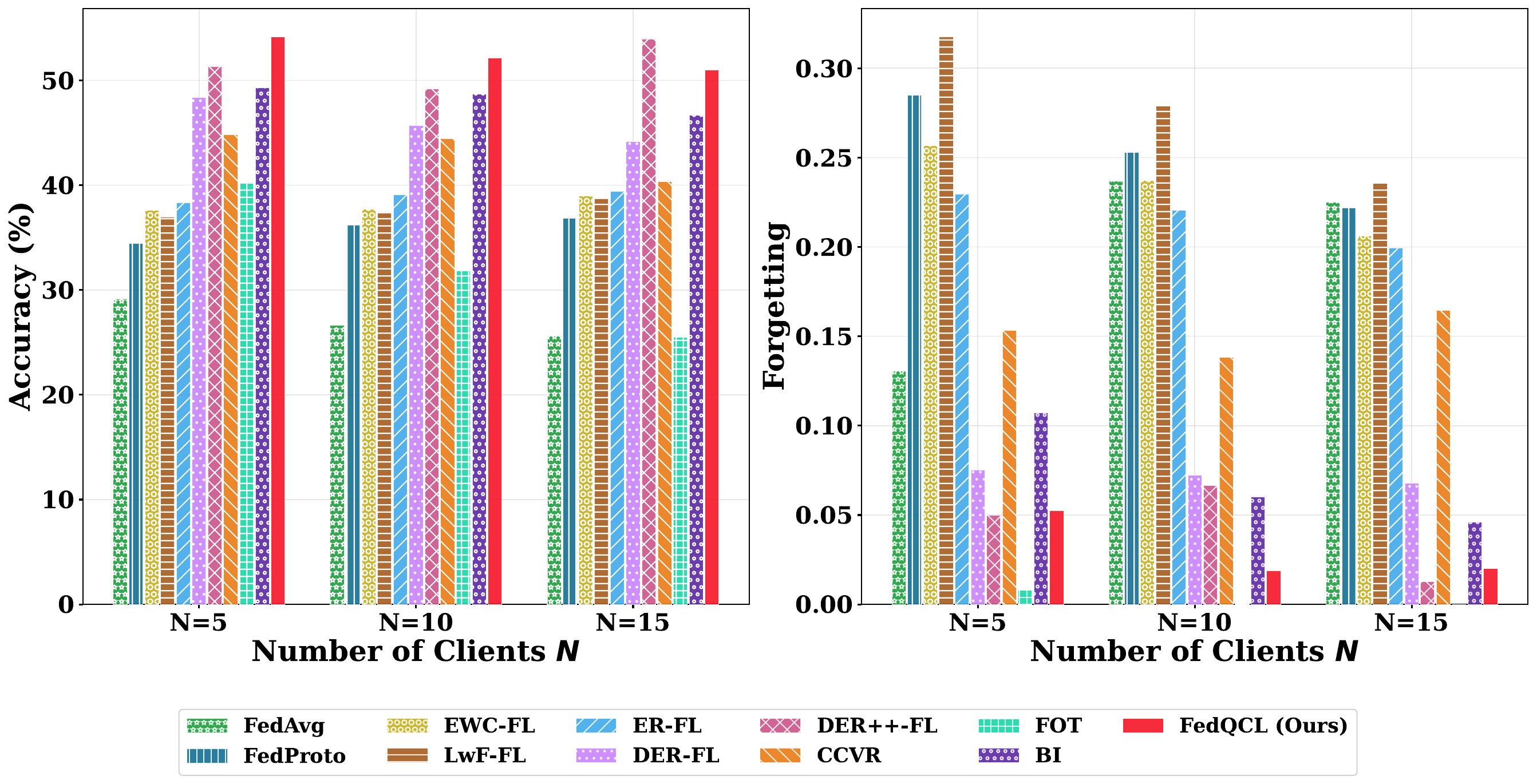}
    \end{subfigure}
    \hfill
    \begin{subfigure}{0.49\linewidth}
        \centering
        \includegraphics[width=\linewidth]{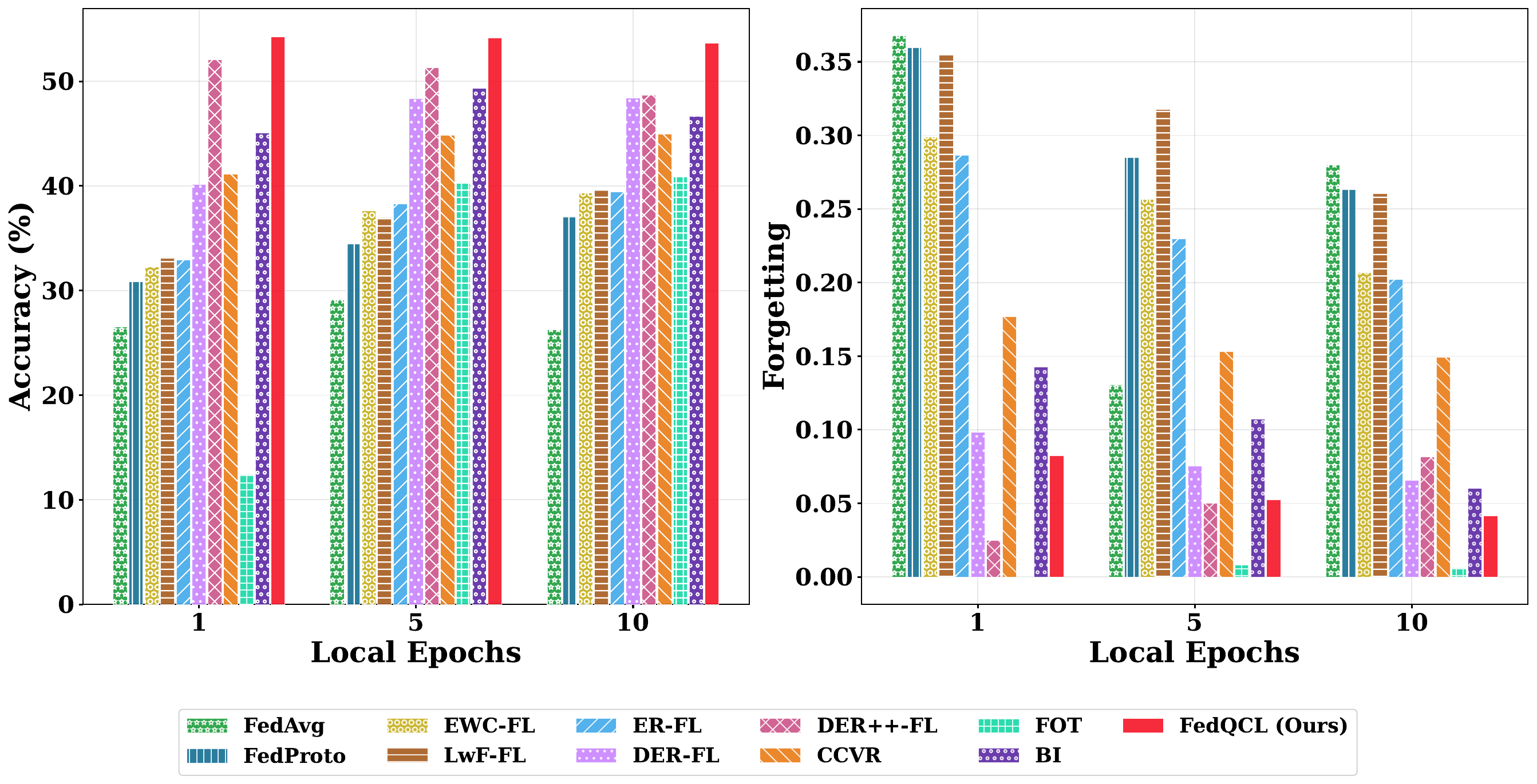}
    \end{subfigure}

    \caption{
        {System-level sensitivity on Split-CIFAR-100.}
        Left: Effect of the number of clients ($N$).
        Right: Effect of local training epochs.
    }
    \label{fig:client_epoch_combined}
\end{figure}

We vary the number of participating clients $N\in\{5,\,10,\,15\}$ on Split-CIFAR-100 to assess the scalability of \textsc{FedQCL}. Results are shown in Figure~\ref{fig:client_epoch_combined} (Left). \textsc{FedQCL} scales gracefully with increasing $N$, maintaining competitive accuracy and low forgetting even at $N{=}15$. Most baselines exhibit some accuracy degradation as $N$ increases, likely because a larger client participation increases data fragmentation per client and further amplifies non-IID effects. \textsc{FedQCL}'s queue-based regularization remains effective at larger federation sizes since the queues are maintained locally and do not introduce any additional communication cost as $N$ grows; each client continues to regulate its own forgetting independently, and the FedAvg aggregation step is unchanged. Forgetting behavior for most methods remains broadly stable across $N$ values, suggesting that federation size primarily affects accuracy rather than catastrophic forgetting in this setting.

\subsubsection{Varying Client Local Epochs}
\label{sec:epoch_sweep}


We vary the number of client local training epochs $R \in \{1, 5, 10\}$ on Split-CIFAR-100, holding the total local update steps per task fixed at $R \times C = 50$. Results are shown in Figure~\ref{fig:client_epoch_combined} (Right). This isolates the effect of synchronization frequency: large $R$ (small $C$) means clients train longer before aggregation, while small $R$ (large $C$) means near-continuous synchronization. Results are shown in Figure~6 (Right). \textsc{FedQCL}'s forgetting decreases monotonically as $R$ increases while accuracy remains stable, indicating the queue-based penalty compensates for reduced aggregation frequency.


 \begin{figure}[htb]
    \centering
    
    \begin{subfigure}{0.305\linewidth}
        \centering
        \raisebox{6mm}{\includegraphics[width=\linewidth]{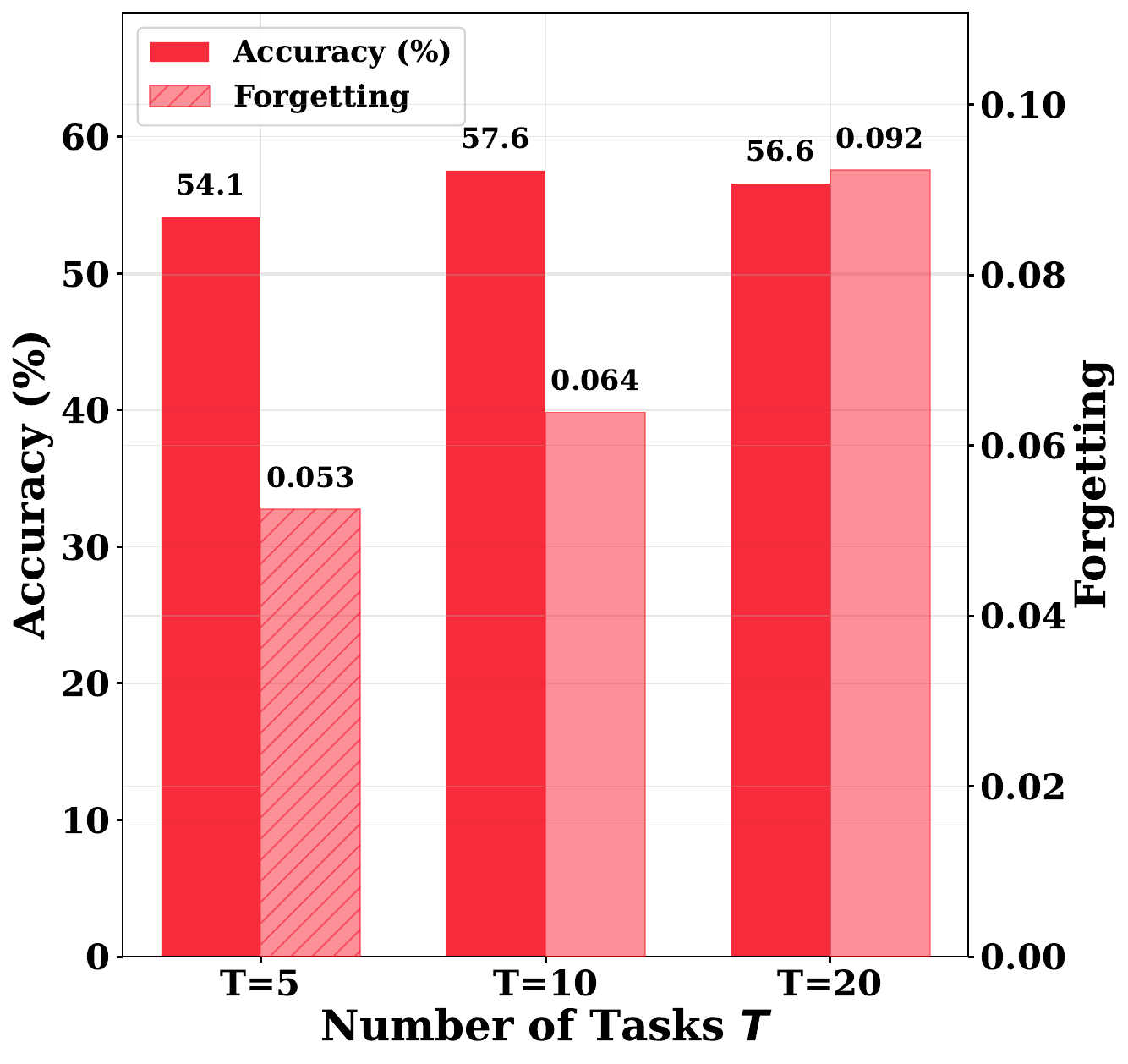}}
    \end{subfigure}
    \hfill
    \begin{subfigure}{0.690\linewidth}
        \centering
        \includegraphics[width=\linewidth]{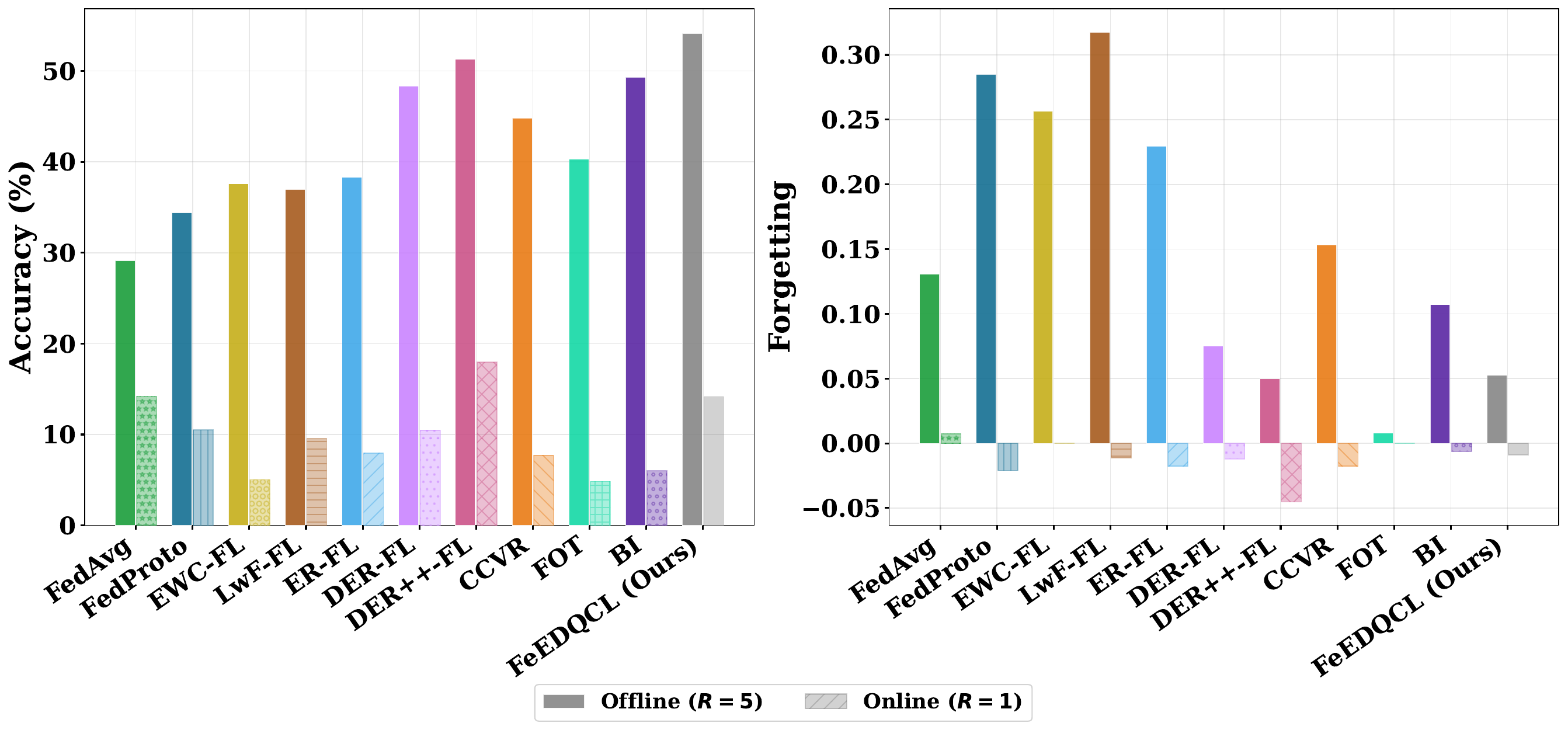}
    \end{subfigure}

    \caption{
        {Scalability and training paradigm analysis on Split-CIFAR-100.}
        Left: Effect of increasing the number of tasks ($T \in \{5,\,10,\,20\}$).
        Right: Comparison between online and offline training settings.
    }
    \label{fig:task_online_combined}
\end{figure}

\subsubsection{Online vs.\ Offline Setting}
\label{sec:online_offline}


We compare the online ($C/T{=}1, R{=}1, B{=}10$, single-pass) and offline ($C/T{=}10,R{=}5,B{>}10$, multi-pass) settings on Split-CIFAR-100. Results are shown in Figure~\ref{fig:task_online_combined} (Right). In the offline setting, \textsc{FedQCL} achieves substantially higher accuracy than its online counterpart, as multiple local epochs allow better convergence on the current task while the queue mechanism simultaneously constrains forgetting. In the online setting, all methods generally suffer reduced accuracy due to limited per-task data exposure. \textsc{FedQCL} retains its relative advantage over baselines in both settings, demonstrating that the queue-based objective is effective regardless of whether data is processed in a single or multiple passes. Forgetting is generally lower in the online setting across most methods, as fewer local gradient steps reduce local drift away from previously learned representations, a trend consistent with the findings in Section~\ref{sec:epoch_sweep}.

\subsubsection{Varying Number of Tasks}
\label{sec:task_sweep}

We evaluate \textsc{FedQCL} on Split-CIFAR-100 with an increasing number of tasks $T\in\{5,\,10,\,20\}$. Results are shown in Figure~\ref{fig:task_online_combined} (Left). Accuracy remains stable across task horizons. The slight improvement at $T{=}10$ relative to $T{=}5$ can be attributed to the finer task granularity: with more but smaller tasks, each task contains fewer classes, which may reduce within-task interference and allow the model to specialize more effectively per task. Forgetting increases modestly from $T{=}5$ to  $T{=}20$, as more tasks expand the queue state that must be jointly managed. Critically, this growth is sublinear, and forgetting remains well-controlled across all settings, demonstrating that \textsc{FedQCL} scales gracefully with the continual learning horizon.

\begin{figure}[htb]
    \centering
    
    \begin{subfigure}{0.24\linewidth}
        \centering
        \includegraphics[width=\linewidth]{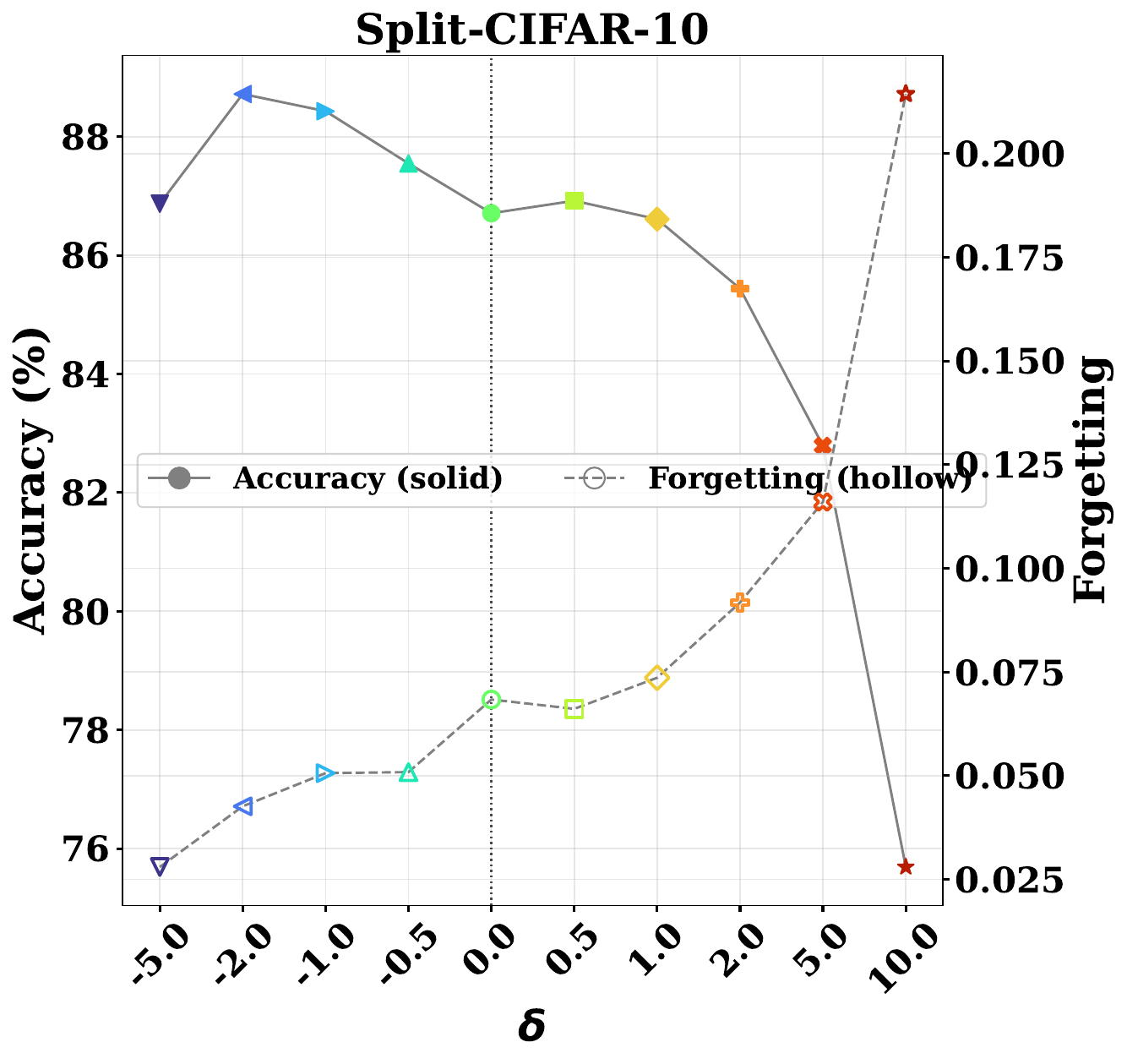}
    \end{subfigure}
    \hfill
    \begin{subfigure}{0.24\linewidth}
        \centering
        \includegraphics[width=\linewidth]{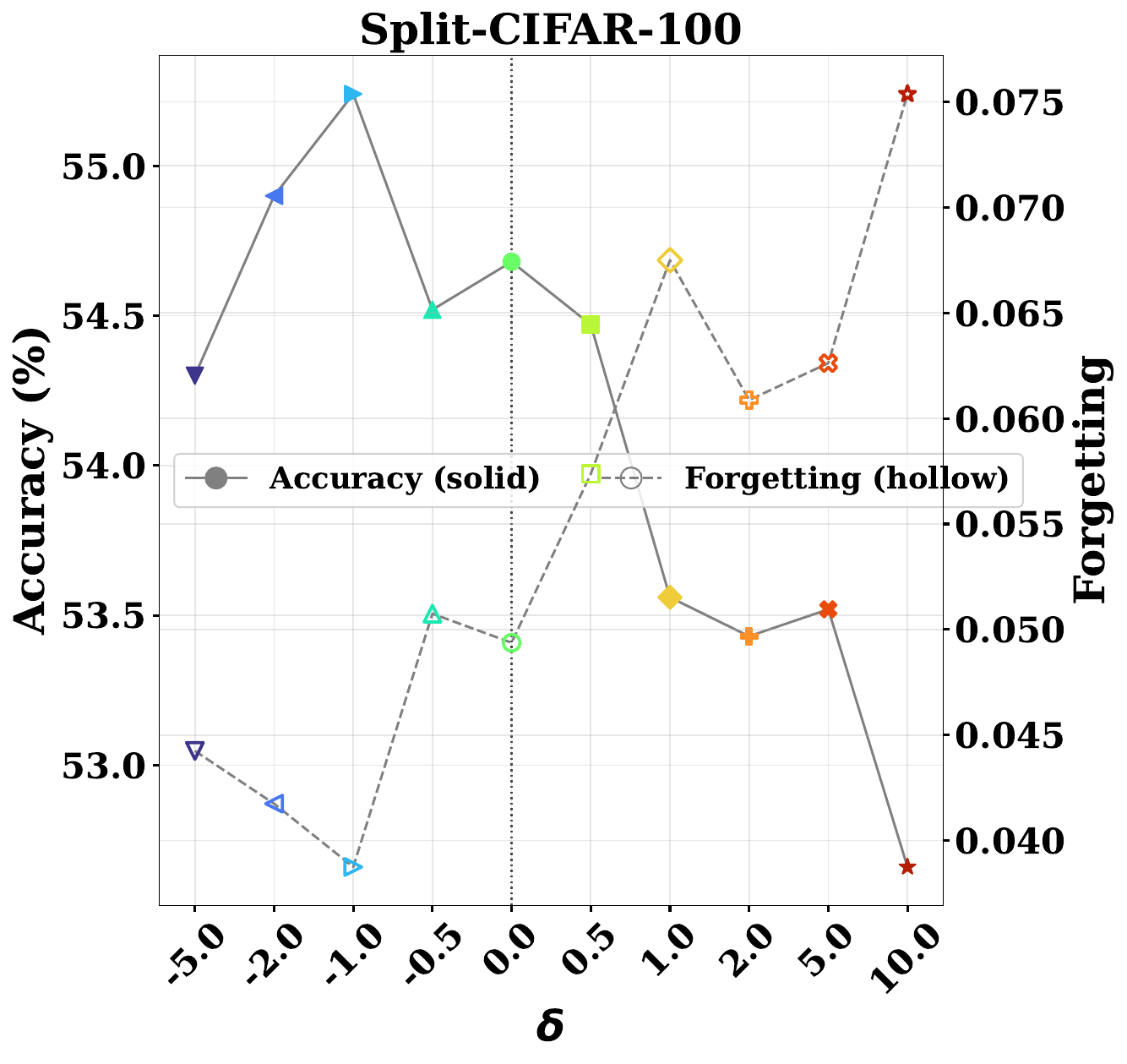}
    \end{subfigure}
    \hfill
    \begin{subfigure}{0.24\linewidth}
        \centering
        \includegraphics[width=\linewidth]{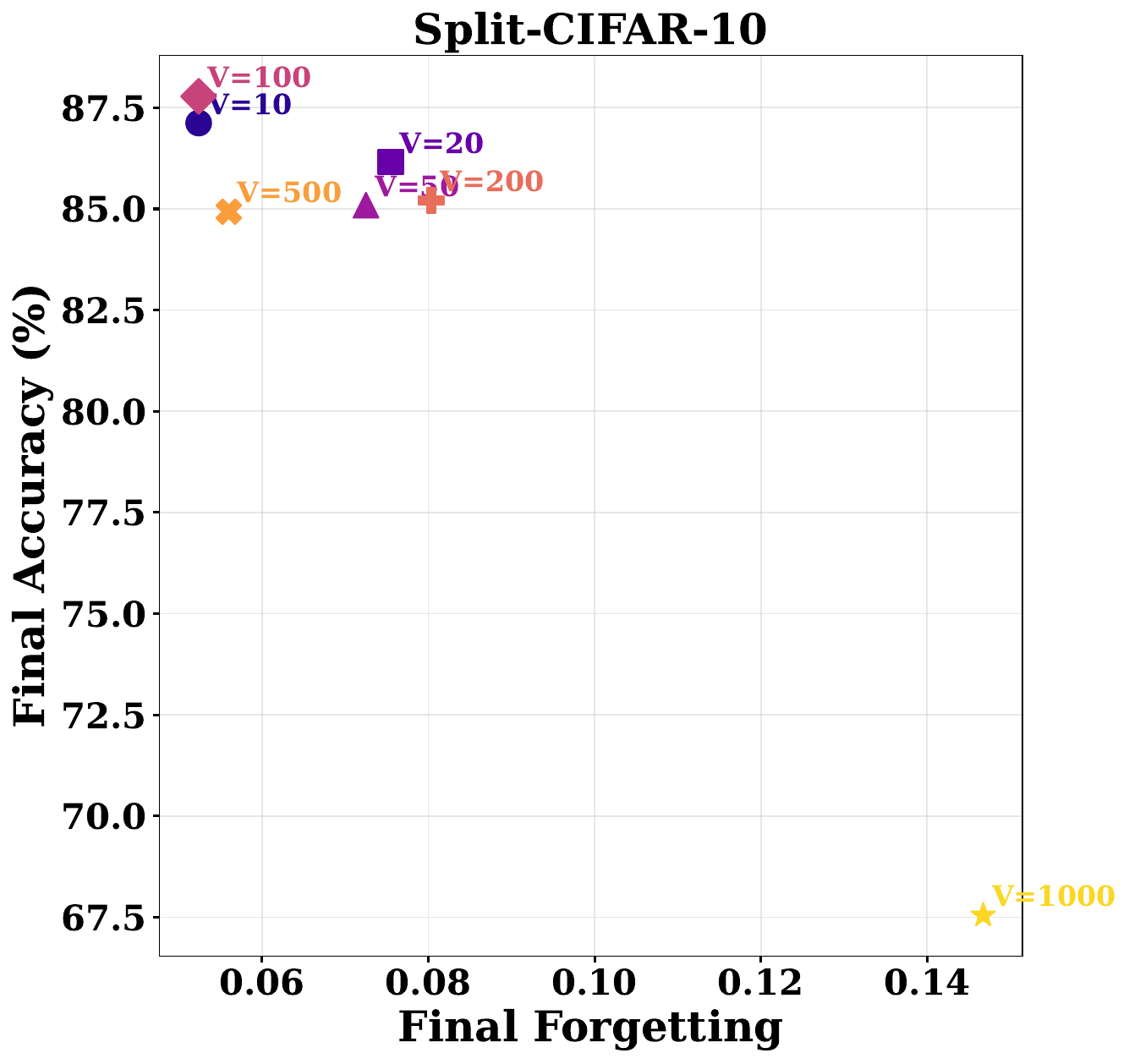}
    \end{subfigure}
    \hfill
    \begin{subfigure}{0.24\linewidth}
        \centering
        \includegraphics[width=\linewidth]{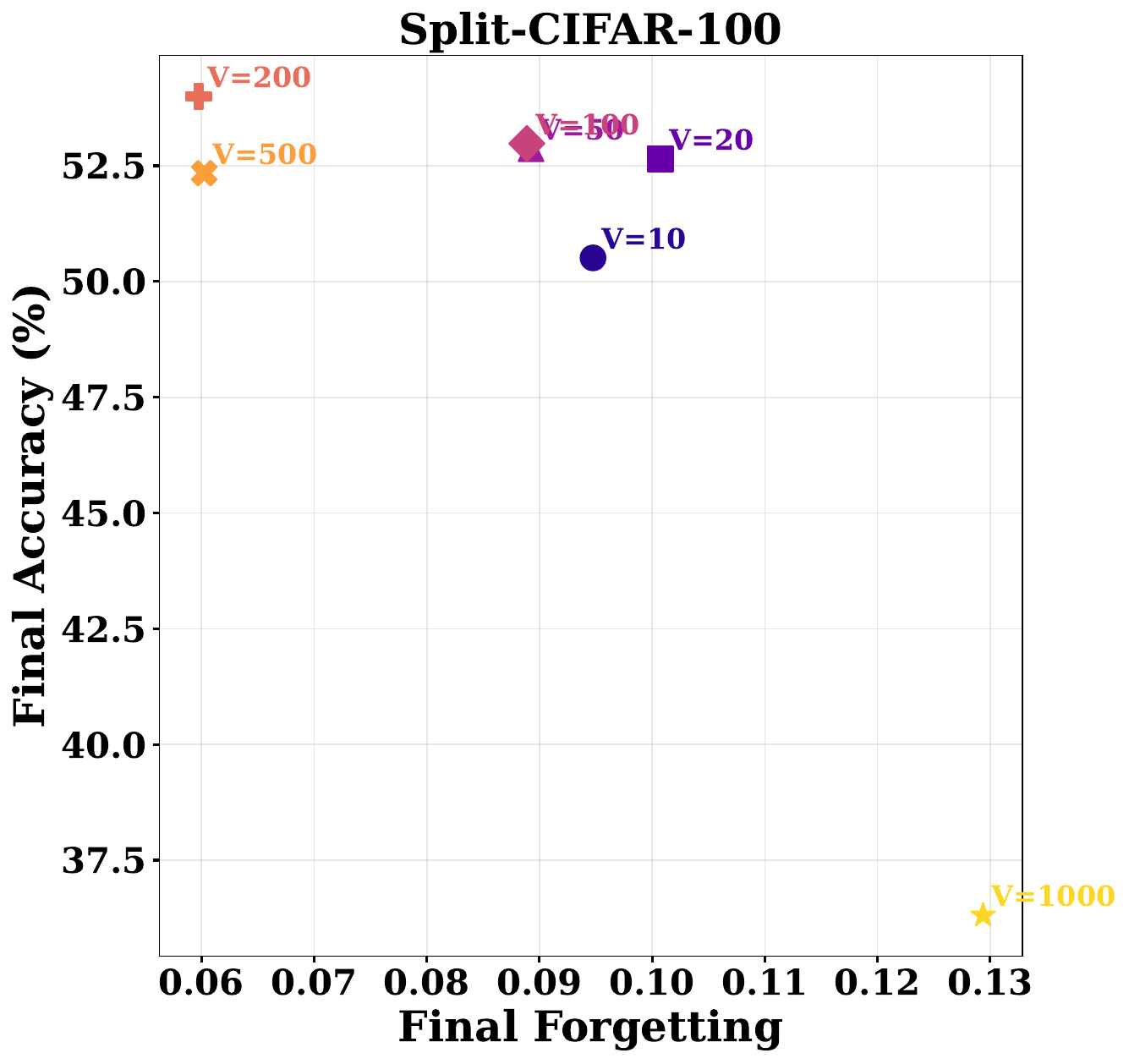}
    \end{subfigure}

    \caption{
        {Additional hyperparameter analysis.}
        Left: Effect of $\delta$.
        Right: Effect of $V$ shown through plasticity–stability Pareto trade-offs.
    }
    \label{fig:supp_hyperparams}
\end{figure}

\subsubsection{Extended Hyperparameter Analysis}
\label{sec:app_hyperparams}

Figure~\ref{fig:supp_hyperparams} provides additional results of the hyperparameter sensitivity analysis presented in Section~\ref{sec:ablation_delta} and~\ref{sec:ablation_V}. The left panels show $\delta$ sweep plots across both datasets. Negative $\delta$ imposes stricter constraints by requiring the replay loss to actively decrease relative to the reference model, leading to larger queue values and stronger regularization pressure. As $\delta$ increases toward positive values, the constraint becomes more lenient: forgetting increases while accuracy can benefit marginally from the additional plasticity. The right panels show plasticity--stability Pareto plots under varying $V$, confirming that moderate $V$ values occupy the upper-left frontier, while very small or very large $V$ push the method toward the lower-left (stability without plasticity) or lower-right (plasticity without stability) corners, respectively.

\begin{figure}[htb]
    \centering
    \includegraphics[width=0.80\linewidth]{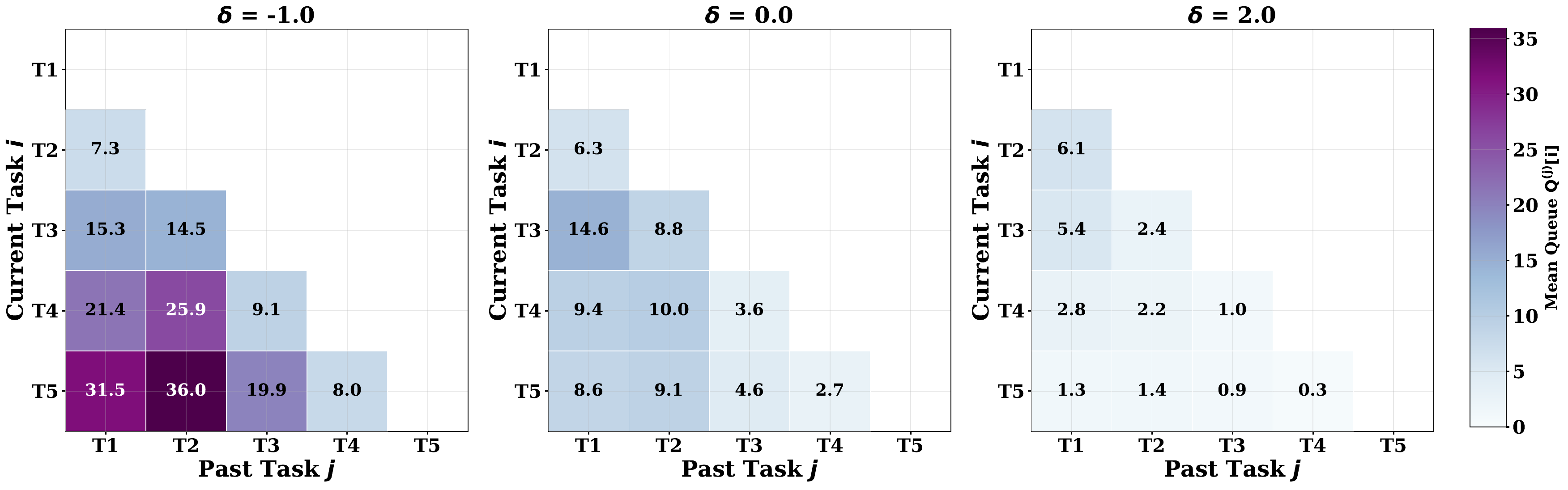}
    \caption{
        {Virtual queue dynamics under varying $\delta$.}
        Each cell $(i,j)$ represents the queue value $Q^{(j)}[i]$ for past task $j$ accumulated during training on task $i$, averaged across clients. Smaller $\delta$ leads to larger queue values, indicating stronger penalization of forgetting.
    }
    \label{fig:queue_heatmap_delta}
\end{figure}

\begin{figure}[htb]
    \centering
    \includegraphics[width=0.80\linewidth]{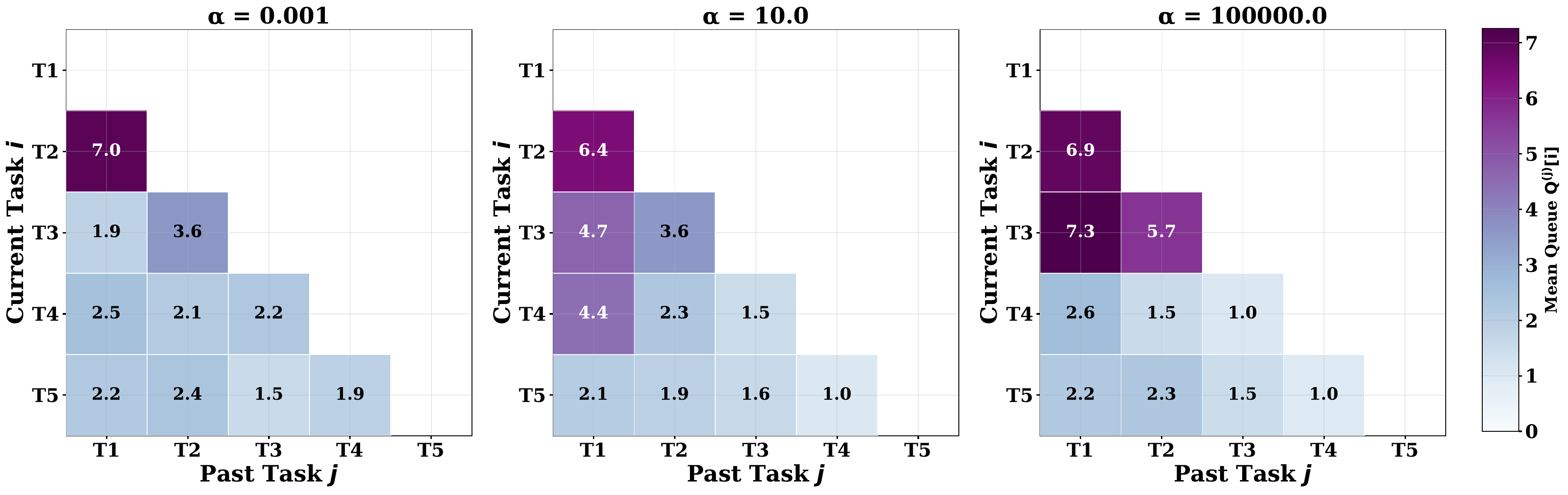}
    \caption{
        {Virtual queue dynamics under varying data heterogeneity $\alpha$.}
        Each cell $(i,j)$ represents the queue value $Q^{(j)}[i]$ for past task $j$ accumulated during training on task $i$, averaged across clients.
    }
    \label{fig:queue_heatmap_alpha}
\end{figure}

\subsubsection{Queue Dynamics: Effect of $\delta$ and $\alpha$}
\label{sec:app_queue_dynamics}

Figure~\ref{fig:queue_heatmap_delta} and~\ref{fig:queue_heatmap_alpha} complements the main-paper analysis (\ref{fig:queue_heatmap_V}) by showing how the virtual queue matrix $Q^{(j)}[i]$ responds to varying forgetting tolerance $\delta$ (top row) and data heterogeneity $\alpha$ (bottom row) on Split-CIFAR-100.

\textbf{Effect of $\delta$ (Figure~\ref{fig:queue_heatmap_delta}):}~Tighter constraints (smaller $\delta$) produce substantially larger queue values throughout the task sequence, particularly for early task pairs. As $\delta$ increases toward more lenient values, queue magnitudes decrease uniformly, confirming that the forgetting constraint is rarely violated when the tolerance is loose. This directly reflects the expected behavior: a tighter budget forces the queue to grow aggressively whenever forgetting exceeds the threshold.

\textbf{Effect of heterogeneity $\alpha$ (Figure~\ref{fig:queue_heatmap_alpha}):}~Queue magnitude does not follow a simple monotonic trend with heterogeneity. This is because the queues reflect the reliability of each client's local replay-loss estimate rather than the severity of global forgetting. Notably, the structural pattern of decreasing queue values along the diagonal is consistent across all $\alpha$ values, suggesting that the queue mechanism reliably captures the forgetting
dynamics irrespective of the heterogeneity level.







\end{document}